\documentclass[final]{cvpr}

\usepackage{times}
\usepackage{epsfig}
\usepackage{graphicx}
\usepackage{amsmath}
\usepackage{amssymb}
\usepackage{algorithm}
\usepackage{algpseudocode}
\usepackage{multirow}
\usepackage{pifont}
\usepackage{booktabs}
\usepackage{epsfig}
\usepackage{booktabs,multirow}
\usepackage{graphics}
\usepackage{threeparttable}
\usepackage{color}
\usepackage[normalem]{ulem}
\usepackage{multirow}
\usepackage{float}
\usepackage{amsfonts}
\usepackage{bm}
\usepackage{rotating}
\usepackage{enumitem}
\usepackage[numbers]{natbib}
\usepackage{array}
\usepackage[table]{xcolor}
\usepackage{colortbl}
\definecolor{myy}{RGB}{126,95,0}
\definecolor{mygray}{gray}{.9}
\definecolor{bblue}{RGB}{30,80,120}
\definecolor{mygray1}{gray}{.7}
\usepackage{bm}
\usepackage{mathtools}
\usepackage{subcaption}
\usepackage{siunitx}
\usepackage[nopar]{lipsum}
\usepackage[export]{adjustbox}

\definecolor{ggray}{RGB}{127,127,127}
\definecolor{mygreen}{RGB}{93,174,86}
\definecolor{myred}{RGB}{192,0,0}

\makeatletter
\newcommand{\thickhline}{%
	\noalign {\ifnum 0=`}\fi \hrule height 1pt
	\futurelet \reserved@a \@xhline
}
\makeatother
\newcommand{\tabincell}[2]{\begin{tabular}{@{}#1@{}}#2\end{tabular}}

\usepackage{caption}
\usepackage[pagebackref=true,breaklinks=true,colorlinks,bookmarks=false]{hyperref}
\usepackage[utf8]{inputenc}

\usepackage{cleveref}
\crefname{section}{§}{§§}
\Crefname{section}{§}{§§}

\begin{document}

\title{RBGNet: Ray-based Grouping for 3D Object Detection}
\author{
Haiyang Wang$^{1}$ ~~~~Shaoshuai Shi$^{2}\thanks{Corresponding author: Shaoshuai Shi.}$~ ~~~~Ze Yang$^{3}$  ~~~~Rongyao Fang$^{4}$  \\
~~~~Qi Qian$^{5}$  ~~~~Hongsheng Li$^{4}$  ~~~~Bernt Schiele$^{2}$  ~~~~Liwei Wang$^{6,7}$   \\
{\normalsize
{$^1$}Center for Data Science, Peking University ~~{$^2$}Max Planck Institute for Informatics}\\
{\normalsize{\hspace*{-16pt}}
~~{$^3$}University of Toronto
~~{$^4$}The Chinese University of Hong Kong
~~{$^5$}Alibaba Group
}
\\
{\normalsize~~{$^6$}Key Laboratory of Machine Perception, MOE, School of Artificial Intelligence, Peking University}\\
{\normalsize~~{$^7$}International Center for Machine Learning Research, Peking University}\\
{\tt\small \{wanghaiyang@stu, wanglw@cis\}.pku.edu.cn
~~\{sshi, schiele\}@mpi-inf.mpg.de}\\
{\tt\small
zeyang@cs.toronto.edu
~~\{rongyaofang@link, hsli@ee\}.cuhk.edu.hk
~~qi.qian@alibaba-inc.com
}
}

\maketitle
\pagestyle{empty}
\thispagestyle{empty}

\begin{abstract}
As a fundamental problem in computer vision, 3D object detection is experiencing rapid growth. To extract the point-wise features from the irregularly and sparsely distributed points, previous methods usually take a feature grouping module to aggregate the point features to an object candidate. However, these methods have not yet leveraged the surface geometry of foreground objects to enhance grouping and 3D box generation. In this paper, we propose the RBGNet framework, a voting-based 3D detector for accurate 3D object detection from point clouds. In order to learn better representations of object shape to enhance cluster features for predicting 3D boxes, we propose a ray-based feature grouping module, which aggregates the point-wise features on object surfaces using a group of determined rays uniformly emitted from cluster centers. 
Considering the fact that foreground points are more meaningful for box estimation, we design a novel foreground biased sampling strategy in downsample process to sample more points on object surfaces and further boost the detection performance. Our model achieves state-of-the-art 3D detection performance on ScanNet V2 and SUN RGB-D with remarkable performance gains. Code will be available at \url{https://github.com/Haiyang-W/RBGNet}.
\end{abstract}
\vspace{-5pt}
\section{Introduction} \label{sec:intro}
3D object detection is becoming an active research topic in computer vision, which aims to estimate oriented 3D bounding boxes and semantic labels of objects in 3D scenes. As a fundamental technique for 3D scene understanding, it plays a critical role in many applications, such as autonomous driving~\cite{bansal2018chauffeurnet, wang2019monocular}, augmented reality~\cite{azuma1997survey,billinghurst2015survey} and domestic robots~\cite{zhu2017target, wang2021collaborative}. 
Unlike the scenarios in the well-studied 2D image problems, 
3D scenes are generally represented by point clouds, a set of unordered, sparse and irregular points captured by depth sensors (\eg, RGB-D cameras, LiDAR sensors), which makes it significantly different from traditional regular input data like images and videos. 
\begin{figure}[t]
  \centering
   \includegraphics[width=0.99\linewidth]{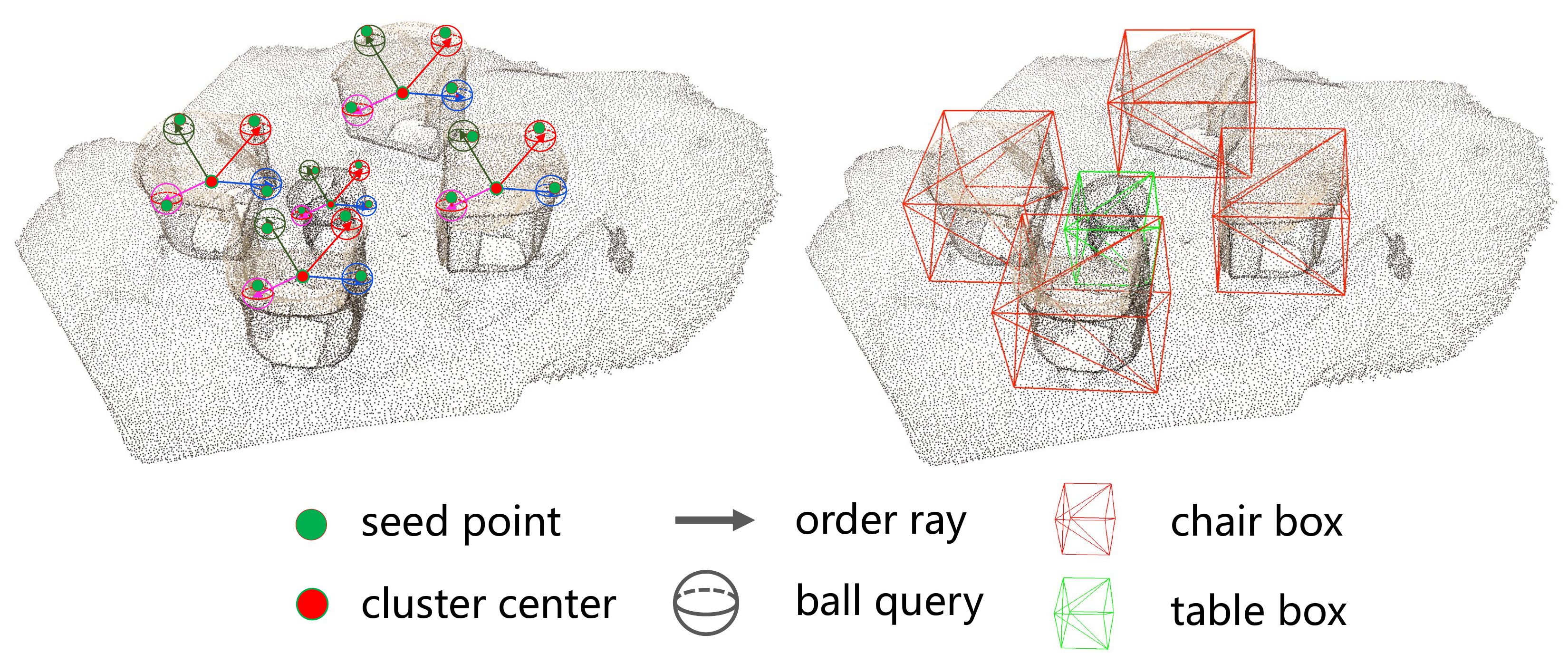}
   \vspace{-6pt}
   \caption{\textbf{3D object detection in point clouds with a ray-based feature grouping module.} Given the point clouds of a 3D scene, our model aggregates the point-wise features on object surface by a group of ordered rays to boost the performance of 3D object detection.}
   \label{fig:fig1}
   \vspace{-12pt}
\end{figure}

Previous 3D detection approaches can be coarsely classified into two lines in terms of point representations, \ie, the grid-based methods and the point-based methods. The grid-based methods generally convert the irregular points to regular data structure such as 3D voxels ~\cite{song2016deep,shi2020points,yan2018second,zhou2018voxelnet,shi2020pv,yin2021cvpr, Deng2021VoxelRT} or 2D bird's eye view maps~\cite{chen2017multi, ku2018joint, liang2019multi, yang2018hdnet, yang2018pixor}. 
Thanks to the great success of PointNet series~\cite{qi2017pointnet,qi2018pointnnetplus}, the point-based methods~\cite{qi2018frustum, shi2019pointrcnn, qi2019deep, zhang2020h3dnet, liu2021group, cheng2021back} directly extract the point-wise features from the irregular and points. These point-wise features are generally enhanced by various feature grouping modules for predicting the 3D bounding boxes. 
However, these feature grouping strategies have not well explored the fine-grained surface geometry to help improve the performance of 3D box generation.

We argue that feature grouping module plays an important role in point-based 3D detectors, and how to better incorporate the foreground object geometry features to enhance the quality of point-wise features is the key to predict better 3D bounding boxes. As shown in Table~\ref{tab:upperbound}, for the popular VoteNet~\cite{qi2019deep} point-wise 3D detector, by simply grouping the features of accurate object surface points to the features of their correct vote centers, the performance can be improved dramatically with a gain of 13.31 on mAP@0.25 for explicit usage of ground truth labels ($2^{nd}$ row of Table~\ref{tab:upperbound}), 
and a gain of 8.79 for implicit usage of ground truth labels ($3^{rd}$ row of Table~\ref{tab:upperbound}).  
Here the ``explicit usage'' indicates that the ground truth labels are not only utilized for grouping the object surface points but also for replacing the vote centers with ground truth centers, while the ``implicate usage'' means the ground truth labels are only used for grouping the object surface points.
These facts inspire us to explore on designing a better feature representation for the surface geometry of foreground objects, to help the prediction of 3D bounding boxes.

\begin{table}
	\centering
	\small
	\resizebox{0.48\textwidth}{!}{
    	\setlength\tabcolsep{2pt}
		\renewcommand\arraystretch{1.0}
		\begin{tabular}{c|c|c||c|c}
			\hline\thickhline
			\rowcolor{mygray}
\tabincell{c}{GT-Features \\(explicit GT-center)}     & \tabincell{c}{GT-Features\\(implicit GT-center)}    & FgSamp     & mAP@0.25       & mAP@0.50       \\\hline\hline
           &            &            & 62.90          & 39.91          \\
\checkmark &            &            & 76.21~(+13.31)  & 53.91~(+14.00)              \\
           & \checkmark &            & 71.69~(+8.79)~~ & 50.71~(+10.80)              \\
		   &            & \checkmark & 71.27~(+8.39)~~ & 50.68~(+10.77)              \\\hline

\hline

	\end{tabular}
	}
	\vspace{-1pt}
	\captionsetup{font=small}
	\caption{\small Results of VoteNet~\cite{qi2019deep} variants on ScannetV2~\cite{dai2017scannet}. 
	\textbf{GT-Features}: Aggregate the features of ground-truth surface points for the 3D box generation of this object, where the ``explicit GT-center'' / ``implicit GT-center'' indicate that the above features are grouped to the ground-truth center / predicted vote centers, respectively.
	\textbf{FgSamp:} 
	FPS is
	only conducted on foreground points.}
	
	\label{tab:upperbound}
	\vspace{-13pt}
\end{table}  

Hence, we present a new 3D detection framework, RBGNet, which is a one-stage 3D detector for 3D object detection from raw point clouds. 
Our RBGNet is built on top of VoteNet~\cite{qi2019deep}, and we propose two novel strategies to boost the performance of 3D object detection by implicitly learning from foreground object features. 

Firstly, we propose the \textit{ray-based feature grouping} that could learn better feature representation of the surface geometry of foreground objects. The learned features are utilized to augment the cluster features for 3D boxes estimation. 
Specifically, we formulate a ray-based mechanism to capture the object surface points, where a number of rays are uniformly emitted from the cluster center with the determined angles (see Fig.~\ref{fig:fig1}). The far bounds of the rays are based on our predicted object scale of this cluster. 
Then a number of anchor points is densely sampled on each ray, where the aggregated local features of each anchor point are utilized to predict whether they are on the object surface to learn the geometry shape. 
Moreover, a coarse-to-fine strategy is proposed to generate different number of anchor points based on the sparsity of different regions. The learned features from all the anchor points will be finally aggregated to boost the features of cluster centers for predicting 3D bounding boxes. 
The experiments (Table~\ref{tab:os_rp}) show that our ray-based feature grouping strategy can effectively encode the surface geometry of foreground objects and significantly improves 3D detection performance.

Secondly, we propose the \textit{foreground biased sampling} strategy to allocate more foreground object points for predicting 3D boxes.
We observe that the points on object surfaces are more useful than those on the background for 3D box estimation (similar observations are also mentioned by \cite{yang20203dssd, shi2020pv}), and $4^{th}$ row of Table~\ref{tab:upperbound} shows that by conducting farthest point sampling only on the ground truth foreground points, the performance of VoteNet~\cite{qi2019deep} could be boosted from 62.90 to 71.27 in terms of mAP@0.25.
Therefore, we propose a simple but effective strategy to sample points biased towards object surface while still keeping the coverage rate of the whole scene. 
Specifically, we append a segmentation head to the point-wise features before each farthest point sampling, where the head will predict the confidence of each point being a foreground point. According to the ranking of their foreground scores, the input points are separated into foreground set and background set. And these two sets will apply farthest point sampling separately, where we sample most target points (\ie, 87.5\% in our case) from the foreground set and a small number (\ie, 12.5\%) from the background set to keep the coverage rate of the whole scene. 
Our foreground biased sampling can produce a more informative sampling of points over foreground objects surface for feature extraction, and the performance gains (Table~\ref{tab:os_rp}) demonstrate its effectiveness.

In a nutshell, our contributions are three-fold: 
1) We propose a novel ray-based feature grouping module to encode object surface points with determined rays, which can learn better surface geometry features of objects to boost the performance of point-based 3D object detectors.
2) We present foreground biased sampling module to focus feature learning of the network on foreground surface points while also keeping the coverage rate for the whole scene, which can incorporate more object points to benefit point-based 3D box generation.
3) Equipped with the above two modules, our proposed RBGNet framework outperforms state-of-the-art methods with remarkable margins both on ScanNetV2~\cite{dai2017scannet} and SUN RGB-D~\cite{sunrgbd}.

\begin{figure*}[t]
  \centering
  \includegraphics[width=0.95\linewidth]{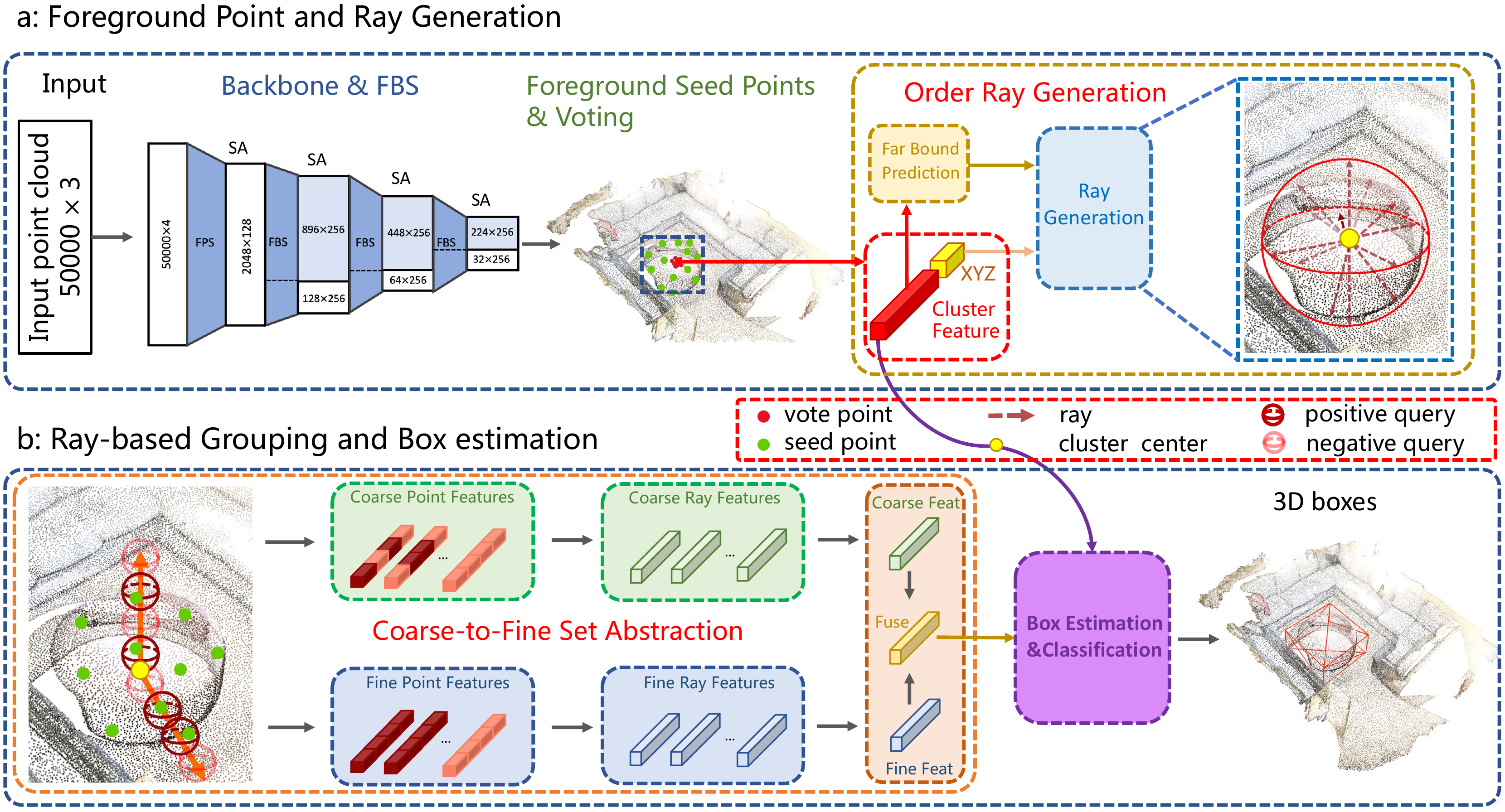}
  \vspace{-8pt}
   \caption{The RBGNet architecture for 3D object detection from point cloud. (a) Generating more foreground seed points and a number of rays emitted from object centers. (b) Object shape encoding by ordered rays and 3D bounding box estimation.}
   \label{fig:overview}
  \vspace{-12pt}
\end{figure*}
\section{Related Work}
\noindent \textbf{3D Object Detection} is challenging due to the irregular, sparse and orderless characteristics of 3D points. Most existing works could be classified into two categories in terms of point cloud representations, \ie, grid-based and point-based. Grid-based approaches transform point clouds to regular data, such as 2D grids~\cite{chen2017multi, zhou2018voxelnet, yang2018pixor} or 3D voxels~\cite{shi2020points,yan2018second,jiang2020pointgroup,shi2020pv,zheng2020cia,zheng2021se,yin2021cvpr,shi2021pv}. 2D grid methods project point clouds to a bird view before proceeding to the rest of the pipeline. Voxel-based methods convert the point clouds into 3D voxels to be processed by 3D CNN or efficient 3D sparse convolution~\cite{graham20183d}, which greatly facilitate 3D object detection. Popularized by PointNet~\cite{qi2017pointnet} and its variants~\cite{qi2018pointnnetplus,jiang2019hierarchical,zhao2019pointweb}, point-based methods~\cite{qi2019deep, zhang2020h3dnet, liu2021group} have become extensively employed on estimating bounding box directly from raw points. Most of existing methods can be considered as a bottom-up manner, which requires point grouping step to obtain object features. Point R-CNN~\cite{shi2019pointrcnn} groups point features within the 3D proposals by point cloud region pooling. VoteNet~\cite{qi2019deep} applies Hough Voting to group the points that vote to the similar center region. Group-free~\cite{liu2021group} implicitly groups point features by an attention module. Although these methods have explored various feature grouping strategies, they have not leveraged the surface geometry of foreground objects. We propose a novel ray-based feature grouping module to encode object shape distribution with determined rays, and the learned features are used to further boost 3D detection performance.

\noindent \textbf{2D Shape Representation.} Shape representation~\cite{xie2020mlcvnet, peng2020deep, xu2019explicit, acuna2018efficient, liang2020polytransform, perreault2021centerpoly} is of particular interest due to the ability to explicitly describe the 2D object shape with points. Polar Mask~\cite{xie2020polarmask} and ESE-SEG~\cite{xu2019explicit} both use polar representation to model the object boundary and then regress the object locations as well as the length of rays emitting uniformly from the object centroids. However, these shape representations may fail to model 3D object surface, because of the limited expressive ability on concave shape, the difficulty of inner center definition. We design a ray-based 3D shape representation to effectively model object surface geometry.
                                                                                     
\noindent \textbf{Point Cloud Sampling.} Sampling~\cite{nezhadarya2020adaptive, lang2020samplenet, oren2018learnsp} aims to represent the original point cloud in a sparse way, plays a key role in point cloud analysis. Farthest point sampling (FPS) has been widely used as a pooling operation~\cite{qi2018pointnnetplus, qi2019deep}, since it can uniformly sample distributed points. However, FPS is agnostic to downstream tasks by a predefined rule, foreground instances with few interior points may lose all points after sampling. 3DSSD~\cite{yang20203dssd} applies a fused FPS based on feature and euclidean distance, but still does not focus on foreground points explicitly. To deal with the dilemma, we design a simple but effective strategy, \textit{foreground biased sampling}, to sample more points on object surface while still keeping the coverage rate of the whole scene.  
\vspace{-1pt}
\section{Methodology}

This section describes the technical details of the proposed RBGNet detector. \S \ref{sec:overview} briefly presents the overview of our approach. Next, \S \ref{sec:ray_method} to \S \ref{sec:learn} elaborate on the network design and the learning objective. 

\subsection{Overview} \label{sec:overview}
RBGNet is a one-stage 3D object detection framework aiming at more accurate bounding box estimation from irregular point clouds. As illustrated in Fig.~\ref{fig:overview}, RBGNet consists of three major components: i) a \textit{backbone network} with foreground biased sampling to extract feature representation from point clouds, ii) a \textit{ray-based feature grouping} module to effectively capture the points on object surface and learn from the shape distribution to augment cluster feature and iii) a \textit{proposal and classification module} followed by 3D non-maximum-suppression (NMS). Our paper mainly focuses on the sampling and grouping modules, so we follow the same proposal and classification strategy as in VoteNet~\cite{qi2019deep} to estimate final bounding boxes. We will describe the technical details in the following parts.

\subsection{Ray-based Feature Grouping} \label{sec:ray_method}
VoteNet~\cite{qi2019deep} has shown tremendous success for 3D object detection. After getting the seed points from the backbone (PointNet++~\cite{qi2018pointnnetplus}), it reformulates traditional Hough voting, and generates object candidates by grouping the seed points whose votes are within the same cluster. The aggregated feature is then used to estimate the 3D bounding boxes and associated semantic labels. However, the quality of the grouping principally determines the reliability of proposal features and detector performance. Some follow-up works ~\cite{chen2020hierarchical, cheng2021back, xie2020mlcvnet} are actually trying to solve this problem, but they have not well explored on the fine-grained surface geometry of foreground objects. To address this limitation, we propose the \textit{ray-based feature grouping} module, which can effectively encode the shape distribution and learn better object features to enhance 3D detection performance.  

\subsubsection{Ray Point Representation.} \label{sec:rpr}
We first illustrate the process of our proposed ray point representation, where two types of anchor points are generated on each ray to encode the object geometry around the cluster centers. These anchor points of all rays will be utilized for the final feature enhancement in \S\ref{sec:feature_enhance}.

\begin{figure}[t]
  \centering
  \includegraphics[width=0.6\linewidth]{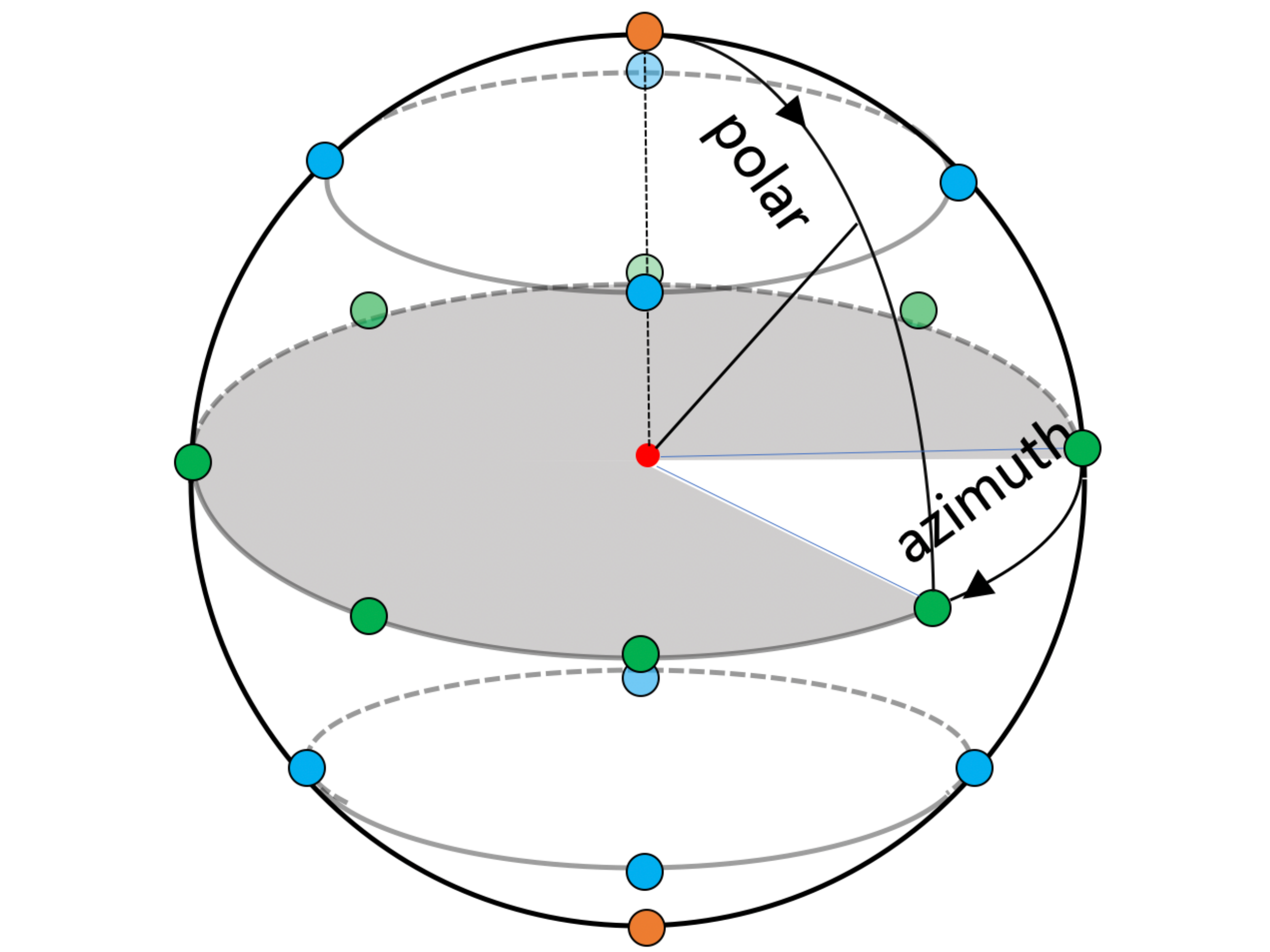}
  \vspace{-6pt}
   \caption{Demonstration of spherical coordinate system and the distribution of 18 rays.}
   \label{fig:polar}
  \vspace{-10pt}
\end{figure}

\noindent\textbf{Formulation of determined rays.}~
We generate a set of vote cluster centers $\{c_i\}_{i=1}^M$ based on the vote sampling and grouping defined in VoteNet~\cite{qi2019deep}, where $c_i = [v_i, f_i]$ indicating the vote center $v_i \in \mathbb{R}^3$ and its features $f_i \in \mathbb{R}^C$, 
$M$ is the number of vote clusters. For each vote cluster, $N$ rays are emitted uniformly from the cluster center  with the determined angles and far bounds. As shown in Fig.~\ref{fig:polar}, the rules for generating rays are based on the spherical coordinate system, which are formulated as follows:
\vspace{-5pt}
\begin{itemize}
	\setlength{\itemsep}{0pt}
	\setlength{\parsep}{-2pt}
	\setlength{\parskip}{-0pt}
	\setlength{\leftmargin}{-10pt}
	\item The polar angle $\theta \in [0, \pi]$ is split into $P$ bins, and each bin corresponds a round surface that is perpendicular to the $Z$-axis. The angle of $p^{th}$ bin is:
		\begin{equation}
   			\small
    		\begin{aligned}
    		\theta_p = \frac{\pi p}{(P ~\text{-}~ 1)} , \quad p \in \{0, ..., P\text{-}1\}.
    		\end{aligned}
    		\label{eq:theta}
		\end{equation}
	\item 
	The number of rays (denoted as $A_{p}$) terminated on the $p^{th}$ round surface is calculated as follows: 
		\begin{equation}
    		\small
    		A_p = \left\{
    		\begin{aligned}
            		&1,~~~~~~~~~~~~~~~~~~~~~~~~~~~~ \text{if~~} p = 0 ~\text{or}~ P~\text{-}~1,\\ 
            		&4 \times p,~~~~~~~~~~~~~~~~~~~~~ \text{if~~} 0 ~\textless~ p \leq \frac{P~\text{-}~1}{2},\\ 
            		&4 \times (P ~\text{-}~ p ~\text{-}~ 1),~~~~~~ \text{if~~} P~\text{-}~1 ~\textgreater~ p ~\textgreater \frac{P~\text{-}~1}{2},\\
    		\end{aligned}
    		\right.\!\!
    		\label{eq:azimuth}
		\end{equation} 
	where $4$ is a hyper-parameter to indicate the factor for the number of sampled rays in each round surface.
	\item With $A_p$ and $\theta_p$, a ray could be determined. Its azimuth angle  $\psi_{p, a} \in [0, 2\pi]$ and polar angle $\theta_{p, a} \in [0, \pi]$  could be formulated as follows:
		\begin{equation}
   			\small
    		\begin{aligned}
    		\psi_{p, a} = \frac{2\pi a}{A_p}, ~~~~\theta_{p, a} = \theta_p, ~~~~a \in \{0, ... A_p~\text{-}~1\}. 
    		\end{aligned}
		\end{equation}
\end{itemize}
Our adopted strategy could generate more uniformly distributed rays to better cover the surrounding region of clusters. 
Given the polar-bin number $P$, the number of rays is $N = \sum_{p=0}^{P-1}A_p$ (\ie $P=9 \rightarrow N=66$ in our case). 
Note that more rays will be generated when the polar angle is closer to $\frac{\pi}{2}$. 

As for the far bounds of the rays of each cluster, all the rays are of the same length as the object scale $l_i$, which is predicted based on the cluster features $f_i$. Here, we explicitly supervise the object scale $l_i$ by regression loss
\begin{equation}
    L_\text{scale-reg} = \frac{1}{I}\sum\limits_{i}\left|\left|l_i - l_i^*\right|\right|_{\eta}\mathbb{I}[\text{$i_{th}$ is positive}],\\
    \label{eq:scale_pre}
    \vspace{-6pt}
\end{equation}
where $l_i^*$ is the half diagonal side of the assigned GT box and $\mathbb{I}[\text{$i_{th}$ is positive}]$ is the indicator function to indicate whether the vote center $c_i$ is around a GT object center (within a radius of $0.3m$). $I$ is the number of positive vote centers. $\eta$ means smooth-$\ell_1$ norm.

\noindent
\textbf{Coarse-to-fine anchor point generation.}~
After generating determined rays, RBGNet samples a number of anchor points 
along each ray. However, it is inefficient to directly sampling points: 1) 
the less important free space and background region 
are still sampled, 2) the number of sampled points in a ball query operation is limited, and thus a lot of object points in dense areas are not captured. 
To address these limitations, we propose a coarse-to-fine anchor point generation strategy. 
It increases the sample efficiency by querying more points for dense areas. 
Inspired by the success of \cite{mildenhall2020nerf}, we adopt a hybrid sampling strategy, which contains two sampling processes: one ``coarse'' and one ``fine''. Before generating anchor points, we first up-sample the seed points back to 2048 points by trilinear interpolation to obtain more meaningful points, especially on object surface. The target point positions for upsampling are the same as the $1^{st}$ SA layers of PointNet++~\cite{qi2018pointnnetplus} backbone.   
For the $i^{th}$ cluster center with cluster features $f$ (we remove the subscript $i$ of $f_i$ for simplicity), we conduct the coarse-to-fine anchor point generation in the following process.  

Firstly, in coarse stage, as for the $n^{th}$ ray, we sample a set of anchor points  as
\begin{align}
	Q_{n}^{(c)}\!= \{q_{n, k}^{(c)} = (x_{n,k}^{(c)}, y_{n,k}^{(c)}, z_{n,k}^{(c)})\}, k \in \{1, \cdots, K_c\},
\end{align} 
where $K_c$ is the number of anchor points sampled on each ray, and the anchor points are generated by stratified sampling to evenly partition the ray into $K_c$ bins. 

To extract local feature of each anchor point, we apply set abstraction ~\cite{qi2018pointnnetplus} to aggregate the features of the seed points around each anchor point. The aggregated local features of anchor point ${q_{n, k}^{(c)}}$ is denoted as  $\rho_{n,k}^{(c)}$. Finally, we append a binary classification module for estimating the positive mask $m_{n,k}^{(c)}$ of point $q_{n, k}^{(c)}$ based on cluster feature $f$ and local feature $\rho_{n,k}^{(c)}$ as follows:
\begin{equation}
    m_{n,k}^{(c)} = \mathcal{F}_{mask}^{(c)}(\rho_{n,k}^{(c)},~ f), 
    \label{eq:psi}
\end{equation}
where the ground-truth of positive masks are calculated by applying ball query operation~\cite{qi2018pointnnetplus} for each anchor point. We assign positive label to an anchor point if some surface points of its assigned GT object are within its ball query region, or the anchor point will be assigned with a negative label.  
Hence this point mask module could predict whether each anchor point belongs to its corresponding object or not. 

Secondly, in fine stage, different from ~\cite{mildenhall2020nerf} which computes sample probability from point density, our fine anchor points are biased towards the dense part of its corresponding object. 
To achieve this goal, 
we apply inverse transform sampling to uniformly generate some $K_f$ anchor points set $Q_{n}^{(f)} = \{q_{n,k}^{(f)}\}_{k=1}^{K_f}$ on positive regions (predicted by the point mask module of coarse stage) of each ray. 
As adopted in the coarse branch, we also extract the local features and predict the positive mask for each fine anchor point. 

Repeat this coarse-to-fine process on all rays, we obtain the coarse and fine local point feature set $\mathcal{P}^{(c)} = \{\rho_{n,k}^{(c)}\}_{k=1,n=1}^{K_c,N}$, $\mathcal{P}^{(f)} = \{\rho_{n,k}^{(f)}\}_{k=1,n=1}^{K_f,N}$, point mask set $\mathcal{M}^{(c)} = \{m_{n,k}^{(c)}\}_{k=1,n=1}^{K_c,N}$, $\mathcal{M}^{(f)} = \{m_{n,k}^{(f)}\}_{k=1,n=1}^{K_f,N}$ and their corresponding positions of anchor points.

\begin{figure}[t]
  \centering
   \includegraphics[width=0.9\linewidth, height=0.55\linewidth]{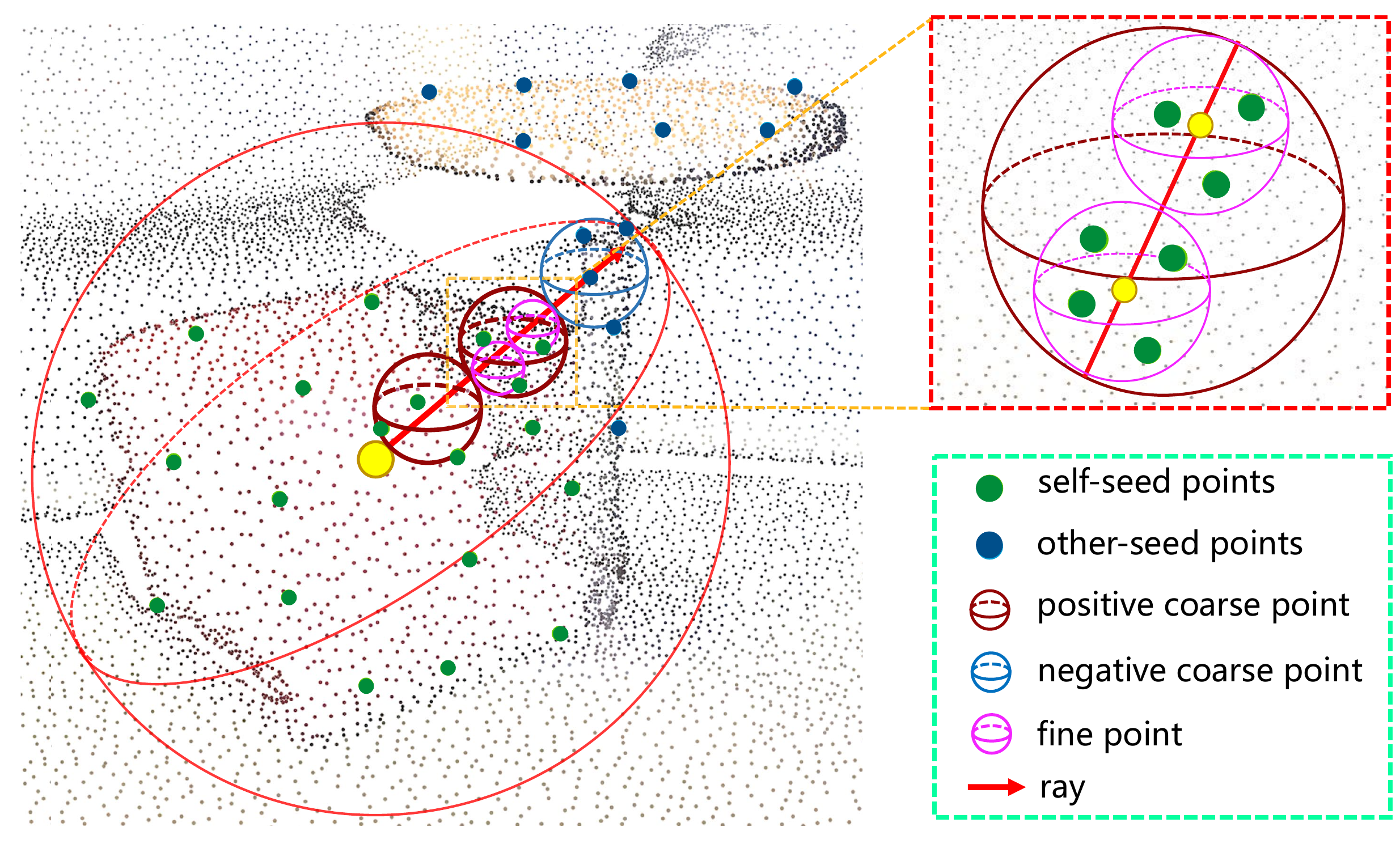}
   \vspace{-10pt}
   \caption{Illustration of coarse-to-fine anchor point generation. Fine sampling is biased towards the dense part of its corresponding object.}
   \label{fig:coarse2fine}
   \vspace{-12pt}
\end{figure}

\vspace{-6pt}
\subsubsection{Feature Enhancement by Determined Rays}\label{sec:feature_enhance}
As discussed in \S \ref{sec:intro}, 
the fined-grained surface geometry of foreground objects plays a crucial role in generating accurate object proposals. 
The process of our proposed coarse-to-fine anchor point generation already encodes such a surface geometry features, since our predicted surface masks and learned local features could implicitly describe the object geometry. 
Here we propose to aggregate those informative features of the anchor points to enhance the quality of cluster features, where the order of rays plays an important roles in the feature aggregation. 

To be specific, given the local features $\mathcal{P}^{(c)}$ and $\mathcal{P}^{(f)}$,  the point masks $\mathcal{M}^{(c)}$ and $\mathcal{M}^{(f)}$ of each anchor point, 
the local features of each anchor points will be masked by setting the features of negative anchor points to zeros. We denote the masked features 
as $\hat{\mathcal{P}}^{(c)}$ and $\hat{\mathcal{P}}^{(f)}$.

To aggregate the learned features orderly based on the determined rays, we formulate a fusion stage to integrate point features in a predefined order of rays. The features of coarse and fine anchor points are aggregated with two separate branches. 
In coarse branch, the masked point features of $n^{th}$ ray, $\{\hat{\rho}_{n,k}^{(c)}\}_{k=1}^{K_c}$, are firstly fused into a single ray feature $r_{n}^{(c)}$. It is implemented by concatenating the features of anchor points in order before being projected to a 32-dimensional features:
\begin{equation}
    \begin{aligned}
    r_{n}^{(c)} = \mathcal{F}_{\text{point}}^{(c)}(\{\hat{\rho}_{n,k}^{(c)}\}_{k=1}^{K_c}, \odot),
    \end{aligned}
    \label{eq:point2ray}
\end{equation}
where $\odot$ means the concatenation operation. 
Then, in the same way, we concatenate all the ray features $\mathcal{R}^{(c)} = \{r_{n}^{(c)}\}_{n=1}^N$ with a determined order and apply a two-layer MLP to generate a 128-dimensional coarse feature: 
\begin{equation}
    \begin{aligned}
    \mu^{(c)} = \mathcal{F}_{\text{ray}}^{(c)}(\{r_{n}^{(c)}\}_{n=1}^N, \odot).
    \end{aligned}
    \label{eq:ray2coarse}
\end{equation}
Note that the predefined order of both anchor points and rays are consistent for each proposal, but different ordering strategies do not affect the performance. 
The fine branch also adopts the same strategy as the coarse branch to generates a 128-dimensional feature $\mu^{(f)}$.
 
Finally, the coarse and fine features are fused as: 
\begin{equation}
    \begin{aligned}
    g = \mathcal{F}_{\text{fuse}}(\mu^{(c)}, \mu^{(f)}).
    \end{aligned}
    \label{eq:fuse}
\end{equation}
The fused feature $g$ is finally combined with the cluster feature $f$ to improve the performance of 3D object detection.

In this way, our RBGNet models the surface geometry implicitly and roughly obtain the size and the position of a possible object in a class-agnostic way, which could greatly benefit the prediction of 3D bounding boxes.  

\subsection{Foreground Biased Sampling}\label{sec:objectsample}
The foreground points provide rich information on predicting their associated object locations and orientations, and force network to capture shape information for more accurate 3D box generation. 
However, the widely-adopted farthest point sampling algorithm in the backbone is agnostic to the downstream tasks and samples a lot of background points. It may bring negative effects for 3D detection. 
Therefore, we design a simple but effective strategy, \textit{Foreground Biased Sampling}, to sample more points on foreground object surfaces while still keeping the coverage rate of the whole scene.

Given the point-wise features encoded by each set abstraction layer, we append a segmentation head for estimating the confidence of each points. The ground-truth segmentation mask is naturally provided by the 3D ground-truth boxes. To be specific, for example, after going through the first SA layer of standard PointNet++~\cite{qi2018pointnnetplus}, we obtain 2048 downsample point set $\mathcal{D} = \{d_j\}_{j=1}^{2048}$ with xyz $\varrho_j$ and 128-dimensional feature $\nu_j$. Then the segmentation head scores each point to be a foreground point or not as: 
\begin{equation}
    \begin{aligned}
    \varepsilon_j = \mathcal{F}^{\text{fore}}(\nu_{j}, \varrho_j) \in [0,1].
    \end{aligned}
    \label{eq:fore_score}
\end{equation}
We sort the confidence scores, select top $\kappa$ to form a foreground set $\mathcal{D}^{(f)} = \{d_j^{(f)}\}_{j=1}^{\kappa}$ and the rest are the background set $\mathcal{D}^{(b)} =\{d_j^{(b)}\}_{j=1}^{2048\text{-}\kappa}$. Due to the concentration of high score points, there is a trade-off between the recall of foreground points and the sampling coverage for the whole scene. Based on this observation, we apply farthest point sampling on foreground and background set separately, and combine them into the final sample set as follows:  
\begin{equation}
    \small
    \begin{aligned}
    \mathcal{D}^{\hat{(f)}} = \text{FPS}(\mathcal{D}^{(f)}), \mathcal{D}^{\hat{(b)}} = \text{FPS}(\mathcal{D}^{(b)}), \mathcal{S} = \mathcal{D}^{\hat{(f)}} \oplus \mathcal{D}^{\hat{(b)}}
    \end{aligned}
    \label{eq:fpsforset}
\end{equation}
where $\mathcal{D}^{\hat{(f)}} = \{d_j^{\hat{(f)}}\}_{j=1}^{\alpha}$ and $\mathcal{D}^{\hat{(b)}} = \{d_j^{\hat{(b)}}\}_{j=1}^{\beta}$, $\alpha$ and $\beta$ are the sample number of foreground and background set. 
$\mathcal{S}$ is the set of final sampled points, which contains more object points while still keeping the coverage rate of the whole scene.
In our case, we sample most target points,(\ie, 87.5\%) from the foreground set and a small number (\ie, 12.5\%) from background set. For example, in downsample process of the $2^{th}$ SA layer (2048 $\rightarrow$ 1024), $\kappa$, $\alpha$ and $\beta$ are 1024, 896 and 128, respectively. 
 
We adopt the cross entropy loss for foreground segmentation. In inference, the confidence score is obtained by the margin between positive class and negative class. 

\subsection{Learning Objective}\label{sec:learn}
The loss function consists of foreground biased sampling $\mathcal{L}_\text{fbs}$, voting regression $\mathcal{L}_{\text{vote-reg}}$, ray-based feature grouping $\mathcal{L}_\text{rbfg}$, objectness $\mathcal{L}_\text{obj-cls}$, bounding box estimation $\mathcal{L}_\text{box}$, and semantic classification $\mathcal{L}_\text{sem-cls}$ losses.
\vspace{-10pt}

\begin{equation}
    \begin{aligned}
    L = \lambda_\text{vote-reg}\mathcal{L}_{\text{vote-reg}} + \lambda_\text{fbs}\mathcal{L}_\text{fbs} + \lambda_\text{rbfg}\mathcal{L}_\text{rbfg} + \\\lambda_\text{obj-cls}\mathcal{L}_\text{obj-cls} + \lambda_\text{box}\mathcal{L}_\text{box} + \lambda_\text{sem-cls}\mathcal{L}_\text{sem-cls}.
    \end{aligned}
\end{equation}
Following the setting in VoteNet~\cite{qi2019deep}, we use the same label assignment and loss terms $\mathcal{L}_{\text{vote-reg}}$, $\mathcal{L}_\text{obj-cls}$, $\mathcal{L}_\text{box}$ and $\mathcal{L}_\text{sem-cls}$. 
$\mathcal{L}_\text{fbs}$ is a cross entropy loss used to supervise foreground sampling (see \S\ref{sec:objectsample}). 
$\mathcal{L}_\text{rbfg}$ is the sum loss of ray-based feature grouping module defined as follows:

\vspace{-10pt}
\begin{equation}
    \begin{aligned}
    \mathcal{L}_\text{rbfg} = \lambda_\text{scale-reg}\mathcal{L}_{\text{scale-reg}} +
    	 \lambda_\text{c-cls}\mathcal{L}_{\text{c-cls}} + \lambda_\text{f-cls}\mathcal{L}_{\text{f-cls}}.
    \end{aligned}
    \label{eq:loss}
\end{equation}
As defined in \S\ref{sec:ray_method}, $\mathcal{L}_{\text{scale-reg}}$ is a smooth $\ell_1$ loss, to explicitly supervise object scale of each proposal. $\mathcal{L}_{\text{c-cls}}$ and $\mathcal{L}_{\text{f-cls}}$ are both cross entropy losses, to supervise our model for querying valid point on each object surface. The detailed balancing factors are in Appendix.

\section{Experiments}
\subsection{Datasets and Evaluation Metric} \label{sec:data_metric}
We evaluate our method on two large-scale indoor 3D scene datasets, \ie, ScanNet V2 ~\cite{dai2017scannet} and SUN RGB-D~\cite{sunrgbd}, and we follow the standard data splits~\cite{qi2019deep} for both of them.

\noindent \textbf{SUN RGB-D}~\cite{sunrgbd} is a single-view RGB-D dataset for 3D scene understanding, which consists of ~5K RGB-D training images annotated with the oriented 3D bounding boxes and the semantic labels for 10 categories. Following the standard data processing in ~\cite{qi2019deep}, we convert the depth images to point clouds using the provided camera parameters. 

\noindent \textbf{ScanNet V2}~\cite{dai2017scannet} consists of richly-annotated 3D reconstructions of indoor scenes. It consists of 1513 training samples (reconstructed meshes converted to point clouds) with axis-aligned bounding box labels for 18 object categories. Compared to SUN RGB-D, its scenes are larger and more complete with more objects. We sample point clouds from the reconstructed meshes by following~\cite{qi2019deep}.

For both datasets, the evaluation follows the same protocol as in VoteNet ~\cite{qi2019deep} using mean average precision(mAP) under different IoU thresholds, \ie, 0.25 and 0.5.

\subsection{Implementation Details.} \label{sec:imple_detail}
\noindent \textbf{Network Architecture Details.} For each 3D scene in the training set, we subsample 50000 points from the scene point cloud as the inputs. For the backbone and voting layers, we follow the same network structure of ~\cite{qi2019deep}, but replace FPS with our proposed Foreground Biased Sampling (FBS) in $2^{nd}\rightarrow 4^{th}$ SA layers. More network details about other parts are given in Appendix. 

\noindent \textbf{Training Scheme.} Our network is end-to-end optimized by using the AdamW optimizer with the batch size 8 per-GPU and initial learning rate of 0.006 for ScanNet V2 and 0.004 for SUN RGB-D. We train the network for 360 epochs on both datasets, and the initial learning rate is decayed by 10x at the 240-th epoch and the 330-th epoch. The gradnorm clip~\cite{zhang2020_b282d173, jin2021nonconvex} is applied to stabilize the training dynamics.\\

\begin{table}
    \centering
    \small
    \resizebox{0.49\textwidth}{!}{
        \setlength\tabcolsep{6pt}
		\renewcommand\arraystretch{1.0}
        \begin{tabular}{c||c|c|c}
            \hline\thickhline
            \rowcolor{mygray}
\rowcolor{mygray}
ScanNet V2 & Backbone & mAP@0.25 & mAP@0.5\\
\hline\hline
F-PointNet~\cite{qi2018frustum}*   & PointNet  & 19.8 & 10.8 \\
3D-SIS~\cite{hou20193d}*       & 3D ResNet & 40.2 & 22.5 \\\hline

HGNet~\cite{chen2020hierarchical}        & GU-net      & 61.3 & 34.4 \\
VoteNet~\cite{qi2019deep}$\dagger$      & PointNet++  & 62.9 & 39.9 \\
3D-MPA~\cite{engelmann20203d}       & MinkNet     & 64.2 & 49.2 \\
H3DNet~\cite{zhang2020h3dnet}       & PointNet++  & 64.4 & 43.4 \\
3Detr~\cite{misra2021end}        & PointNet++    & 65.0 & 47.0    \\
BRNet~\cite{cheng2021back}        & PointNet++  & 66.1 & 50.9 \\
VENet~\cite{xie2021venet}        & PointNet++  & 67.7 & - \\
Group-free~\cite{liu2021group}   & PointNet++  & 67.2(66.6) & 49.7(49.0) \\
Our(R66, O256)          & PointNet++  & \textbf{70.2(69.6)} & \textbf{54.2(53.6)} \\
\hline\hline
H3DNet~\cite{zhang2020h3dnet}       & 4$\times$PointNet++  & 67.2 & 48.1 \\
Group-free~\cite{liu2021group}   & PointNet++w2$\times$  & 69.1(68.6) & 52.8(51.8) \\
Our(R66, O512)          & PointNet++w2$\times$  & \textbf{70.6(69.9)} & \textbf{55.2(54.7)} \\
\hline

\end{tabular}
}
	\caption{Performance comparison on the ScanNetV2~\cite{dai2017scannet} val set with state-of-the-art. The main comparison is based on the best results of multiple experiments, the number within the bracket is the average result of 25 trials. $\dagger$: reported by~\cite{liu2021group}, which is better than the official paper. *: use RGB as addition inputs.}
\label{tab:scannet}
\end{table}

\begin{table}
    \centering
    \small
    \resizebox{0.49\textwidth}{!}{
        \setlength\tabcolsep{6pt}
		\renewcommand\arraystretch{1.0}
        \begin{tabular}{c||c|c|c}
            \hline\thickhline
            \rowcolor{mygray}
\rowcolor{mygray}
SUN RGB-D & Backbone & mAP@0.25 & mAP@0.5\\
\hline\hline
F-PointNet~\cite{qi2018frustum}*   & PointNet   & 54.0 & - \\
ImVoteNet~\cite{qi2020imvotenet}*    & PointNet++  & 63.4 & - \\
MTC-RCNN~\cite{park2021multi}*    & PointNet++  & 65.3 (64.7) & 48.6 (48.2) \\
\hline
3Detr~\cite{misra2021end}        & PointNet++    & 59.1 & 32.7    \\
VoteNet~\cite{qi2019deep}$\dagger$     & PointNet++  & 59.1 & 35.8 \\
MLCVNet~\cite{xie2020mlcvnet}      & PointNet++  & 59.8 & -    \\
HGNet~\cite{chen2020hierarchical}        & GU-net      & 61.6 & - \\
H3DNet~\cite{zhang2020h3dnet}       & 4$\times$PointNet++  & 60.1 & 39.0 \\
BRNet~\cite{cheng2021back}        & PointNet++  & 61.1 & 43.7 \\
VENet~\cite{xie2021venet}        & PointNet++  & 62.5 & 39.2 \\
Group-free~\cite{liu2021group}   & PointNet++ & 63.0(62.6) & 45.2(44.4) \\
Our(R66, O256)    & PointNet++  & \textbf{64.1(63.6)} & \textbf{47.2(46.3)} \\
\hline

\end{tabular}
}
	\caption{3D object detection results on the SUN-RGB-D~\cite{sunrgbd} val set. The main comparison is also based on the best results of multiple experiments between different methods. $\dagger$: reported by~\cite{liu2021group}, which is better than the official paper. *: use RGB as addition inputs and our method is geometric only.}
\label{tab:sunrgbd}
\vspace{-10pt}
\end{table}

\vspace{-10pt}
\subsection{Comparison with state-of-the-art methods.} \label{sec:compare}
For performance benchmarking, we compare with a wide range state-of-the-art methods on ScanNet V2 and SUN RGB-D. We follow the previous work \cite{liu2021group} and also report both best results and average results.\\
\noindent \textbf{ScanNet V2.} The results are summarized in Table \ref{tab:scannet}. With the same backbone network of a standard PointNet++, our approach achieves 70.2 mAP@0.25 and 54.2 mAP@0.5 using 66 rays and 256 object candidates, which is 2.5 and 3.3 better than previous best methods~\cite{xie2021venet, cheng2021back} using the same backbones. With stronger backbones and more sampled object candidates just like ~\cite{liu2021group}, \ie, $2\times$ more channels and 512 candidates, our approach is also improved dramatically, achieving 70.6 mAP@0.25 and 55.2 mAP@0.5, which is still 1.5 and 2.4 better than ~\cite{liu2021group}. Notably, we also only use geometric input (point cloud) as previous works did.\\
\noindent \textbf{SUN RGB-D.} We also evaluate our RBGNet against several competing approaches on SUN RGB-D dataset, which is also a standard benchmark for 3D object detection.  The results are summarized in Table \ref{tab:sunrgbd}. Our base model with standard PointNet++ achieves 64.1 on mAP@0.25 and 47.2 on mAP@0.5, which outperforms all previous state-of-the-arts point-only methods.

\subsection{Ablation Studies and Discussions} \label{sec:ab}
\begin{table}
	\centering
	\resizebox{0.36\textwidth}{!}{
    	\setlength\tabcolsep{6pt}
		\renewcommand\arraystretch{1.0}
		\begin{tabular}{c|c|c|c}
			\hline\thickhline
			\rowcolor{mygray}
FBS & Ray Feat~(66) & mAP@0.25    &   mAP@0.5   \\\hline\hline
           &                 &  66.2 &   48.2     \\
\checkmark &                 &  67.1 &   49.0     \\
 & \checkmark                 &  69.0 & 52.9      \\
\checkmark & \checkmark      &   \textbf{69.6} &   \textbf{53.6}     \\
\hline
\hline

	\end{tabular}
	}
	\vspace{-1pt}
	\caption{Effect of ray-based feature grouping module and foreground biased sampling.
	}
	\label{tab:os_rp}
	\vspace{-6pt}
\end{table}

\begin{figure*}[t]
  \centering
  \vspace{-5pt}
  \includegraphics[width=0.8\linewidth]{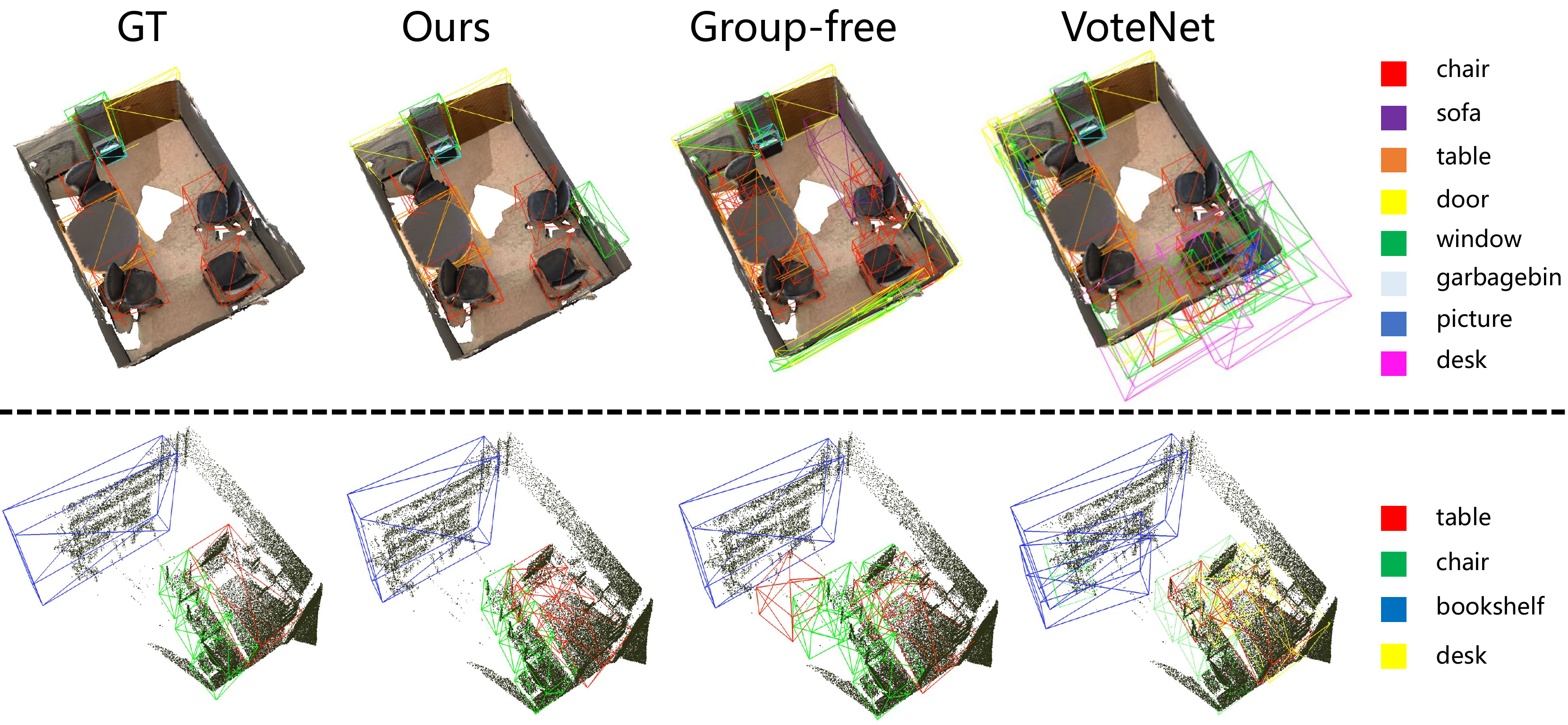}
  \vspace{-5pt}
   \caption{Qualitative results on ScanNet V2 (top) and SUN RGB-D (bottom). The baseline methods are Group-free~\cite{liu2021group} and VoteNet~\cite{qi2019deep}. Our method can generate high-quality and compact bounding boxes compared with other methods.}
   \label{fig:demo}
    \vspace{-8pt}
   
\end{figure*}

\begin{table}
	\centering
	\small
	\vspace{-2pt}
	\resizebox{0.45\textwidth}{!}{
    	\setlength\tabcolsep{6pt}
		\renewcommand\arraystretch{1.0}
		\begin{tabular}{c|c|c|c}
			\hline\thickhline
			\rowcolor{mygray}
number of ray  & objp recall & mAP@0.25    &   mAP@0.5   \\\hline\hline
0          &  - &   67.1          &   49.0     \\
6          &  38.1  &   68.4          &   51.6     \\
18         &  63.3  &   68.7          &   52.0     \\
38         &  75.8  &   69.2          &   52.7     \\
66         &  78.1  &   69.6          &   53.6     \\
102        &  86.5   &  \textbf{69.9}          &   \textbf{53.9}     \\
\hline
\hline

	\end{tabular}
	}
	\caption{Ablation study on the performance of ray-base feature grouping module with different ray number.
	}
	\label{tab:diffray}
	\vspace{-8pt}
	
\end{table}

\begin{table}
	\centering
	\resizebox{0.37\textwidth}{!}{
    	\setlength\tabcolsep{6pt}
		\renewcommand\arraystretch{1.0}
		\begin{tabular}{c|c|c}
			\hline\thickhline
			\rowcolor{mygray}
method          & mAP@0.25 & mAP@0.5\\\hline\hline
Voting~\cite{qi2019deep}          & 67.1  & 49.0\\
RoI-Pooling~\cite{shi2019pointrcnn}    & 67.6  & 49.9     \\
Back-tracing~\cite{cheng2021back}    & 67.7  & 50.1     \\
Group-free~\cite{liu2021group}    & 68.1  & 50.5     \\
Our(R66)             & \textbf{69.6}  & \textbf{53.6}       \\

\hline
\hline

	\end{tabular}
	}
	\caption{Comparison with other grouping-based methods.
	}
	\label{tab:group_ab}
	\vspace{-8pt}
\end{table}

\begin{table}
	\centering
	\small
	\resizebox{0.49\textwidth}{!}{
    	\setlength\tabcolsep{6pt}
		\renewcommand\arraystretch{1.0}
		\begin{tabular}{c|c||c|c|c}
			\hline\thickhline
			\rowcolor{mygray}
method               & dataset           &   $2^{nd}$(1024)  & $3^{rd}$(512) & $4^{th}$(256)     \\\hline\hline
FPS~\cite{qi2019deep} & ScanNet V2     &  31.1     & 30.8       & 30.3           \\
F-FPS~\cite{yang20203dssd} & ScanNet V2     &  40.4      &42.1       & 43.8           \\
Ours & ScanNet V2      &  \textbf{51.2}        & \textbf{73.2} & \textbf{87.8}           \\\hline
FPS~\cite{qi2019deep} & SUN RGB-D           &  17.9                     & 17.8    & 17.7              \\
F-FPS~\cite{yang20203dssd} & SUN RGB-D     &  21.3            &22.9       & 23.5           \\
Ours & SUN RGB-D            &  \textbf{30.8}                     & \textbf{45.3}    & \textbf{65.1}              \\
\hline
\hline

	\end{tabular}
	}
	\vspace{-1pt}
	\caption{Foreground percentage statistics on PointNet++ of different sampling approaches. 
	}
	\vspace{-15pt}
	\label{tab:statistic_fore}
\end{table}

In this section, a set of ablative studies are conducted on 
ScanNet V2 dataset, to investigate effectiveness of essential components of our algorithm. We follows \cite{liu2021group} and report the average performance of 25 trials by default.

\noindent \textbf{Effect of ray-based representation.} We first ablate the effects of ray-based feature grouping in Table \ref{tab:os_rp}, \ref{tab:diffray} and \ref{tab:group_ab}. As evidenced in the first three rows in Table \ref{tab:os_rp}, with ray-based feature grouping, our model performs better, \ie, 66.2 $\rightarrow$ 69.0, 48.2 $\rightarrow$ 52.9 on mAP@0.25 and mAP@0.5. Note that our model is implemented based on a strong baseline. The first row in Table \ref{tab:os_rp} is actually the VoteNet~\cite{qi2019deep} with corner loss regularization, \textit{vote} sampling in vote aggregation layer and optimized hyper-parameters. Even on such a strong baseline (almost close to state-of-the-art already, 66.2 vs 66.6~\cite{liu2021group} on mAP@0.25), ray-based feature grouping module still boosts our model with a remarkable improvement. 

Our ray-based feature grouping module also works well in a wide range of hyper-parameters, such as the number of rays. Table \ref{tab:diffray} shows its performance with different ray number. More rays can bring significant performance improvement, especially in the mAP@0.5. Compared with the setting without any rays, our ray-102 model performs much better on mAP@0.25 and mAP@0.5 by 2.8 and 4.9, respectively. 
For the recall of object points, 
the second column shows that more rays can capture surface points more completely. 
Considering the trade-off between memory usage and performance improvement, our model finally adopts the variant with 66 rays though more rays is better.

\begin{table}
	\centering
	\resizebox{0.45\textwidth}{!}{
    	\setlength\tabcolsep{6pt}
		\renewcommand\arraystretch{1.0}
		\begin{tabular}{c|c|c|c}
			\hline\thickhline
			\rowcolor{mygray}
method         & mAP@0.25 & mAP@0.5 & frames/s \\\hline\hline
MLCVNet~\cite{xie2020mlcvnet} & 64.5 & 41.4 & 5.37 \\
BRNet~\cite{cheng2021back}     & 66.1  & 50.9   & \textbf{7.37} \\
H3DNet~\cite{zhang2020h3dnet}    & 67.2  & 48.1  & 3.75  \\
Group-free~\cite{liu2021group}    & 67.3  & 48.9  &  6.64  \\\hline\hline
Our(R6)            & 69.0  & 52.3   & 7.23    \\
Our(R18)            & 69.0  & 52.6   & 5.70    \\
Our(R38)            & 69.7  & 53.3   & 5.27    \\
Our(R66)            & \textbf{70.2}  & \textbf{54.2}   & 4.75    \\

\hline
\hline

	\end{tabular}
	}
	\caption{Comparison on realistic inference speed on ScanNet V2. Note that mAP@0.25 and 0.5 are the best results of multiple experiments. 
	}
	\label{tab:time}
	\vspace{-10pt}
	
\end{table}

To further demonstrate the effectiveness of ray-based feature grouping module, we refer several grouping strategies in 3D object detection as baselines and compare with them. For a fair comparison, we only switch the feature aggregation mechanism while all other settings remain unchanged. Table \ref{tab:group_ab} shows that our approach achieves more reliable detection results than others with a remarkable margin (1.5 on mAP@0.25 and 3.1 on mAP@0.5). \\
\noindent \textbf{Effect of foreground biased sampling.} Table \ref{tab:os_rp} also demonstrates the effectiveness of the foreground biased sampling strategy. We can observe that, 
it improves the performance in both settings with and without feature grouping module. This verifies the necessity of sampling more foreground points for 3D object detection tasks. To further ablate the effectiveness of FBS, we compare the foreground points recall of $2^{nd}\rightarrow4^{th}$ SA layers among different sub-sampling methods in Table \ref{tab:statistic_fore}. Our sampling strategy draws better performance with a large margin.
\vspace{-3pt}
\subsection{Inference Speed.}\label{sec:ins}~
The realistic inference speed of our method is competitive with other state-of-the-art methods. For a fair comparison, all experiments are run on the same workstation (single NVIDIA Tesla V100 GPU, 256G RAM, and Xeon E5-2650 v3). The results are shown in Table.~\ref{tab:time}. Our method achieves better performance with a competitive speed.

\vspace{-5pt}
\section{Conclusion}
In this paper, we have presented the RBGNet, a novel framework for 3D object detection from point clouds. We propose the ray-base feature grouping module, which can encode object surface geometry with determined rays and learn better geometric features to boost the performance of point-based 3D detectors. We also introduce the foreground biased sampling to sample more points on object surface while keeping the coverage rate for the whole scene. All of the above designs enable our model to achieve state-of-the-art performance on ScanNet V2 and SUN RGB-D benchmarks with remarkable performance gains. 

\noindent \textbf{Acknowledgments.} Liwei Wang was supported by National Key R$\&$D Program of China (2018YFB1402600), BJNSF (L172037) and Alibaba Group through Alibaba Innovative Research Program. Project 2020BD006 supported by PKUBaidu Fund.

\newpage 

{\small

\bibliographystyle{ieee_fullname}
}

\clearpage
\appendix
In the supplementary material, we first elaborate on fine sampling (\S\ref{sec:finesamp}). Then, more implementation details and ray-based representation discussion are provided in \S\ref{sec:netloss} and \S\ref{sec:ray_discuss}. Finally, we present per-category evaluation, visualization of positive anchor points and quantitative results from \S\ref{sec:perclass} to \S\ref{sec:qresults}. We also discuss the limitation of RBGNet in \S\ref{sec:discuss}.

\section{Details on Fine Sampling} \label{sec:finesamp}
In this section, we will provide more technical details on fine sampling. After obtaining coarse point masks, $\mathcal{M}^{(c)} = \{m_{n,k}^{(c)}\}_{k=1,n=1}^{K_c,N}$, we generate fine anchor points biased towards the dense part of its corresponding object. To achieve this goal, we apply inverse transform sampling to uniformly generate $K_f$ anchor points set $Q^{(f)} = \{q_{k}^{(f)}\}_{k=1}^{K_f}$ on positive regions of the $n^{th}$ ray (we remove the subscript $n$ of $Q_n^{(f)}$ for simplicity) based on the predicted coarse point masks $\{m_{k}^{(c)}\}_{k=1}^{K_c}$.

To be specific, we first normalize the coarse point masks as $\hat{m}_{k}^{(c)} = m_{k}^{(c)}/\sum_{j=1}^{K_c}m_{j}^{(c)}$ to produce a piecewise-constant probability density function (PDF). Then we translate it into the cumulative distribution function (CDF). Finally, sampling with the CDF at uniform steps concentrates samples around regions with positive coarse point masks. 

\begin{figure}[t]
  \centering
   \includegraphics[width=0.99\linewidth]{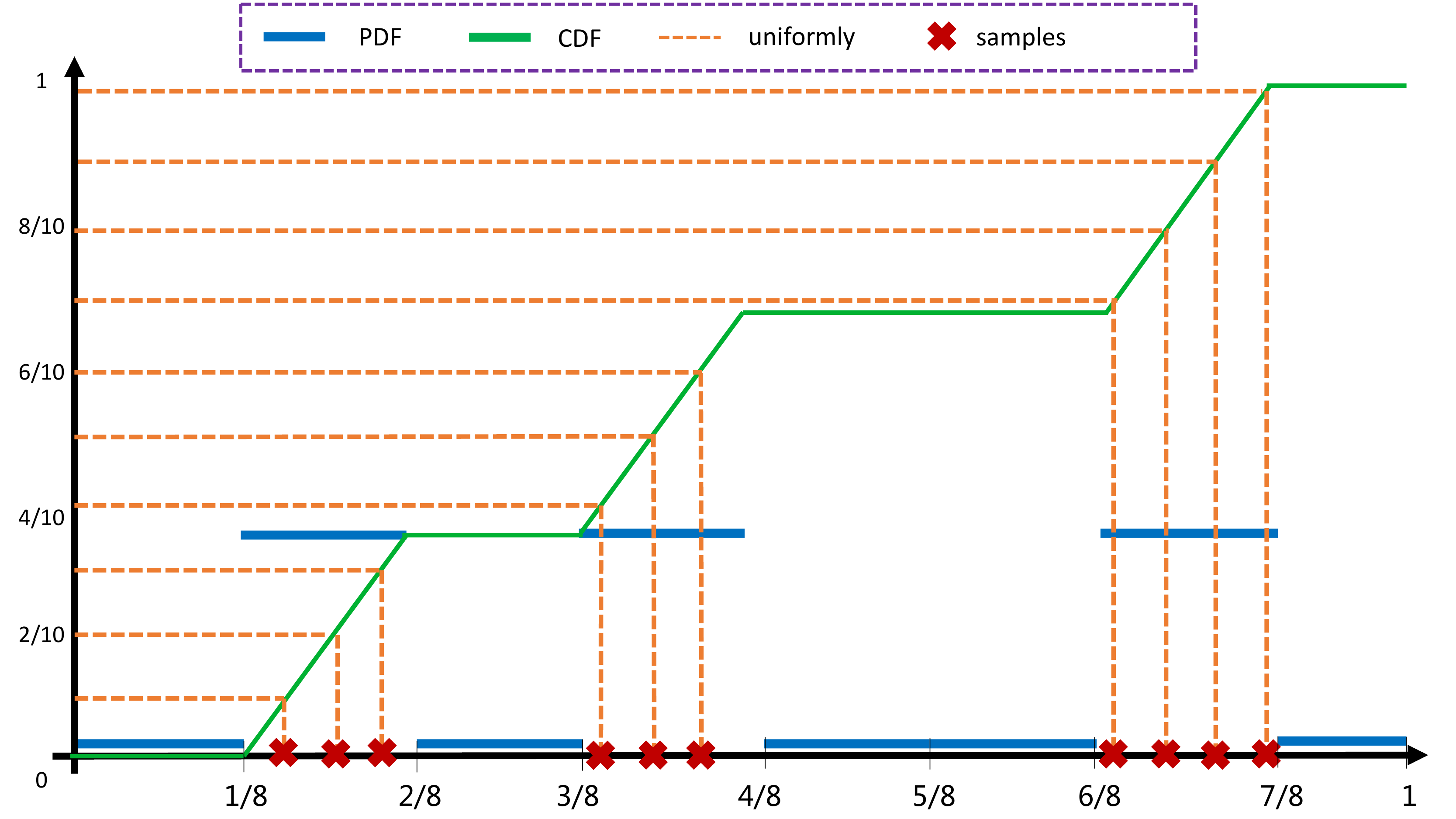}
   \caption{One case for visualization of fine sampling.}
   \label{fig:inversesamp}
\end{figure}

To further illustrate it, we provide a demo case and visualize it in Fig.\ref{fig:inversesamp}. The number of coarse and fine anchor points on each ray is 8 and 10 respectively.
\begin{itemize}
	\setlength{\itemsep}{0pt}
	\setlength{\parsep}{-2pt}
	\setlength{\parskip}{-0pt}
	\setlength{\leftmargin}{-10pt}
	
	\item The predicted coarse point masks of $n^{th}$ ray are:
	\begin{equation}
	    \small
    	\begin{aligned}
    	\{m_{k}^{(c)}\}_{k=1}^{K_c} = \{0, 1, 0, 1, 0, 0, 1, 0\}	.
    	\end{aligned}
    	\label{eq:coarsemask}
	\end{equation}
	\item We compute the piece-wise PDF by normalizing $m_{k}^{(c)}$:
	\begin{equation}
	    \small
    	\begin{aligned}
    	\{\hat{m}_{k}^{(c)}\}_{k=1}^{K_c} = \{0., 1/3, 0., 1/3, 0., 0., 1/3, 0.\}.
    	\end{aligned}
    	\label{eq:coarsemask}
	\end{equation}
	\item Then we convert PDF to CDF, $\{C_{k}^{(c)}\}_{k=1}^{K_c}$
	\begin{equation}
		\small
    	\begin{aligned}
    	\{C_{k}^{(c)}\}_{k=1}^{K_c} = \{0., 1/3, 1/3, 2/3, 2/3, 2/3, 1., 0.\}.
    	\end{aligned}
    	\label{eq:coarsemask}
	\end{equation}
	\item Finally, sample 10 points based on the CDF at uniform steps and inverse them to original distribution. As demonstration in Fig.\ref{fig:inversesamp}, the relative distance of fine points from object center are as follows:
	\begin{equation}
	    \small
    	\begin{aligned}
    	\{\hat{q}_{k}^{(f)}\}_{k=1}^{K_f} = \{0.1625, 0.2000, 0.2375, 0.4000,\\
    	                             0.4375, 0.4750, 0.7625, 0.8000,\\
    	                              0.8375, 0.875\} \times RayScale	.
    	\end{aligned}
    	\label{eq:coarsemask}
	\end{equation}

\end{itemize}

\section{Implementation Details.} \label{sec:netloss}
\subsection{RBGNet architecture details.}~As mentioned in the main paper, the RBGNet architecture consists of a backbone with foreground biased sampling, a voting layer, a ray-based feature grouping module and a proposal module.

\begin{table}
    \centering
    \resizebox{0.49\textwidth}{!}{
		\renewcommand\arraystretch{1.0}
        \begin{tabular}{c||c|c|c|c|c}
            \hline\thickhline
            \rowcolor{mygray}
\rowcolor{mygray}
 layer name    & input layer & $\kappa$ & $\alpha$ & $\beta$ & MLP Channels \\\hline
sa2            & sa1         & 1024     & 896      & 128     & [256, 256] \\
sa3            & sa2         & 512     & 448      & 64     & [256, 256] \\
sa4            & sa3         & 256     & 224      & 32     & [256, 256] \\
\hline\hline

\end{tabular}
}
	\caption{Backbone network architecture: FBS parameters.}
\label{tab:fbs_para}
\end{table}  

\begin{table*}
	\centering
	\resizebox{1.0\textwidth}{!}{
		\setlength\tabcolsep{6pt}
		\renewcommand\arraystretch{1.0}
		\begin{tabular}{c||c|c|c|c|c|c|c|c|c|c|c|c|c|c|c|c|c|c||c}
			\hline\thickhline
			\rowcolor{mygray}
			\rowcolor{mygray}
			& cab & bed & chair & sofa & tabl & door & wind & bkshf & pic & cntr & desk & curt & frig & showr & toil & sink & bath & ofurn & mAP\\\hline
			VoteNet~\cite{qi2019deep} & 47.87 & 90.79 & 90.07 & \textbf{90.78} & 60.22 & 53.83 & 43.71 & 55.56 & 12.38 & 66.85 & 66.02 & 52.37 & 52.05 & 63.94 & 97.40 & 52.32 & 92.57 & 43.37 & 62.90\\
			MLCVNet~\cite{xie2020mlcvnet} & 42.45 & 88.48 & 89.98 & 87.40 & 63.50 & 56.93 & 46.98 & \textbf{56.94} & 11.94 & 63.94 & 76.05 & 56.72 & \textbf{60.86} & 65.91 & 98.33 & 59.18 & 87.22 & 47.89 & 64.48\\ 
			BRNet~\cite{cheng2021back} & 49.90 & 88.30 & 91.90 & 86.90 & 69.30 & 59.20 & 45.90 & 52.10 & 15.30 & 72.00 & 76.80 & 57.10 & 60.40 & 73.60 & 93.80 & 58.80 & 92.20 & 47.10 & 66.10\\
			H3DNet*~\cite{zhang2020h3dnet} & 49.40 & 88.60 & 91.80 & 90.20 & 64.90 & \textbf{61.00} & 51.90 & 54.90 & 18.60 & 62.00 & 75.90 & 57.30 & 57.20 & 75.30 & 97.90 & 67.40 & 92.50 & 53.60 & 67.20\\
			Group-free~\cite{liu2021group} & \textbf{55.40} & 86.60 & 91.80 & 86.60 & 73.30 & 54.50 & 49.40 & 47.70 & 13.10 & 63.30 & 82.40 & 63.30 & 53.20 & 74.00 & 99.20 & 67.70 & 91.70 & 55.80 & 67.20\\
			\hline\hline
			Ours & 52.62 & \textbf{91.34} & \textbf{93.07} & 89.71 & \textbf{73.57} & 60.10 & \textbf{51.96} & 53.53 & \textbf{20.01} & \textbf{72.65} & \textbf{82.57} & \textbf{63.58} & 59.79 & \textbf{76.03} & \textbf{99.28} & \textbf{74.79} & \textbf{92.67} & \textbf{55.88} & \textbf{70.20}\\
			
			\hline
			
		\end{tabular}
	}
	\caption{3D object detection scores per category on the ScanNetV2 dataset, evaluated with mAP@0.25 IoU. * means that H3DNet~\cite{zhang2020h3dnet} only provide the checkpoint with 4 PointNet++ backbones.}
	\label{tab:scan0.25}
\end{table*}  

\begin{table*}
	\centering
	\resizebox{1.0\textwidth}{!}{
		\setlength\tabcolsep{6pt}
		\renewcommand\arraystretch{1.0}
		\begin{tabular}{c||c|c|c|c|c|c|c|c|c|c|c|c|c|c|c|c|c|c||c}
			\hline\thickhline
			\rowcolor{mygray}
			\rowcolor{mygray}
			& cab & bed & chair & sofa & tabl & door & wind & bkshf & pic & cntr & desk & curt & frig & showr & toil & sink & bath & ofurn & mAP\\\hline
			VoteNet~\cite{qi2019deep} & 14.62 & 77.85 & 73.11 & 80.49 & 46.54 & 25.09 & 15.98 & 41.85 & 2.50 & 22.34 & 33.35 & 25.02 & 31.04 & 17.58 & 87.75 & 23.05 & 81.60 & 18.66 & 39.91\\
			H3DNet*~\cite{zhang2020h3dnet} & 20.50 & 79.70 & 80.10 & 79.60 & 56.20 & 29.00 & 21.30 & 45.50 & 4.20 & 33.50 & 50.60 & 37.30 & 41.40 & 37.00 & 89.10 & 35.10 & 90.20 & 35.40 & 48.10\\
			Group-free~\cite{liu2021group} & 23.80 & 77.20 & 81.60 & 65.10 & 62.80 & 35.00 & 21.30 & 39.40 & 7.00 & 33.10 & \textbf{66.30} & 39.30 & 43.90 & \textbf{47.00} & 91.20 & 38.50 & 85.10 & 37.40 & 49.70\\
			BRNet~\cite{cheng2021back} & 28.70 & 80.60 & 81.90 & 80.60 & 60.80 & 35.50 & 22.20 & 48.00 & 7.50 & 43.70 & 54.80 & 39.10 & \textbf{51.80} & 35.90 & 88.90 & 38.70 & 84.40 & 33.00 & 50.90\\
			
			\hline\hline
			Ours & \textbf{30.69} & \textbf{80.95} & \textbf{86.48} & \textbf{84.82} & \textbf{66.45} & \textbf{40.37} & \textbf{29.59} & \textbf{48.60} & \textbf{7.96} & \textbf{44.76} & 59.14 & \textbf{40.83} & 44.80 & 39.78 & \textbf{92.92} & \textbf{45.30} & \textbf{90.90} & \textbf{41.49} & \textbf{54.21}\\
			
			\hline
			
		\end{tabular}
	}
	\caption{3D object detection scores per category on the ScanNetV2 dataset, evaluated with mAP@0.50 IoU. * means that H3DNet~\cite{zhang2020h3dnet} only provide the checkpoint with 4 PointNet++ backbones.}
	\label{tab:scan0.5}
\end{table*} 

\begin{table}
    \centering
    \resizebox{0.39\textwidth}{!}{
		\renewcommand\arraystretch{1.0}
        \begin{tabular}{c||c|c}
            \hline\thickhline
            \rowcolor{mygray}
\rowcolor{mygray}
 layer name    & Input channel & MLP Channels \\\hline
$\mathcal{F}^{(c)}_{point}$ & 5$\times$32 & [32,] \\
$\mathcal{F}^{(f)}_{point}$ & 3$\times$32 & [32,] \\
$\mathcal{F}^{(c)}_{ray}$ & 66$\times$32 & [256, 128] \\
$\mathcal{F}^{(f)}_{ray}$ & 66$\times$32 & [256, 128] \\
$\mathcal{F}_{fuse}$ & 256 & [256, 128] \\
\hline\hline

\end{tabular}
}
	\caption{Detailed layer parameters of Feature Enhance module. In our case, the number of rays ($N$), coarse points ($K_c$) and fine points ($K_f$) are 66, 5, and 3.}
\label{tab:feh_para}
\end{table}

\begin{table*}
    \centering
    \resizebox{0.9\textwidth}{!}{
        \setlength\tabcolsep{6pt}
		\renewcommand\arraystretch{1.0}
        \begin{tabular}{c||c|c|c|c|c|c|c|c|c|c||c}
            \hline\thickhline
            \rowcolor{mygray}
\rowcolor{mygray}
     & bathtub & bed & bookshelf & chair & desk & dresser & nightstand & sofa & table & toilet & mAP\\\hline
VoteNet~\cite{qi2019deep} & 75.50 & 85.60 & 31.90 & 77.40 & 24.80 & 27.90 & 58.60 & 67.40 & 51.10 & 90.50 & 59.10\\
MLCVNet~\cite{xie2020mlcvnet} & 79.20 & 85.80 & 31.90 & 75.80 & 26.50 & 31.30 & 61.50 & 66.30 & 50.40 & 89.10 & 59.80\\ 
H3DNet*~\cite{zhang2020h3dnet} & 73.80 & 85.60 & 31.00 & 76.70 & 29.60 & 33.40 & 65.50 & 66.50 & 50.80 & 88.20 & 60.10\\
BRNet~\cite{cheng2021back} & 76.20 & 86.90 & 29.70 & 77.40 & 29.60 & 35.90 & 65.90 & 66.40 & 51.80 & 91.30 & 61.10\\
HGNet~\cite{chen2020hierarchical} & 78.00 & 84.50 & \textbf{35.70} & 75.20 & \textbf{34.30} & 37.60 & 61.70 & 65.70 & 51.60 & 91.10 & 61.60\\
Group-free~\cite{liu2021group} & 80.00 & 87.80 & 32.50 & 79.40 & 32.60 & 36.00 & 66.70 & 70.00 & 53.80 & 91.10 &63.00\\
\hline\hline
Ours & \textbf{80.68} & \textbf{88.41} & 34.56 & \textbf{82.79} & 32.09 & \textbf{38.76} & \textbf{66.77} & \textbf{71.06} & \textbf{54.55} & \textbf{91.37} & \textbf{64.10}\\

\hline

\end{tabular}
}
	\caption{3D object detection scores per category on the SUN RGB-D dataset, evaluated with mAP@0.25 IoU. * means that H3DNet~\cite{zhang2020h3dnet} only provide the checkpoint with 4 PointNet++ backbones.}
\label{tab:sun0.25}
\end{table*}  

\begin{table*}
    \centering
    \resizebox{0.9\textwidth}{!}{
        \setlength\tabcolsep{6pt}
		\renewcommand\arraystretch{1.0}
        \begin{tabular}{c||c|c|c|c|c|c|c|c|c|c||c}
            \hline\thickhline
            \rowcolor{mygray}
\rowcolor{mygray}
     & bathtub & bed & bookshelf & chair & desk & dresser & nightstand & sofa & table & toilet & mAP\\\hline
VoteNet~\cite{qi2019deep} & 45.40 & 53.40 & 6.80 & 56.50 & 5.90 & 12.00 & 38.60 & 49.10 & 21.30 & 68.50 & 35.80\\
H3DNet*~\cite{zhang2020h3dnet} & 47.60 & 52.90 & 8.60 & 60.10 & 8.40 & 20.60 & 45.60 & 50.40 & 27.10 & 69.10 & 39.00\\
BRNet~\cite{cheng2021back} & 55.50 & 63.80 & 9.30 & 61.60 & 10.00 & \textbf{27.30} & 53.20 & 56.70 & 28.60 & 70.90 & 43.70\\
Group-free~\cite{liu2021group} & 64.00 & 67.10 & 12.40 & 62.60 & \textbf{14.50} & 21.90 & 49.80 & 58.20 & 29.20 & 72.20 & 45.20\\

\hline\hline
Ours & \textbf{65.74} & \textbf{68.02} & \textbf{12.99} & \textbf{65.46} & 12.81 & 25.84 & \textbf{54.89} & \textbf{59.55} & \textbf{32.38} & \textbf{74.50} & \textbf{47.22}\\

\hline

\end{tabular}
}
	\caption{3D object detection scores per category on the SUN RGB-D dataset, evaluated with mAP@0.50 IoU. * means that H3DNet~\cite{zhang2020h3dnet} only provide the checkpoint with 4 PointNet++ backbones.}
\label{tab:sun0.5}
\end{table*}  
The backbone network, based on the PointNet++ architecture, has four set abstraction layers and two feature up-sampling layers. We follow the same layer parameters (\eg ball-region radius, number of sample points and MLP channels) as VoteNet~\cite{qi2019deep}. To sample points biased towards object surface, we append a segmentation head for estimating the foreground confidence of each point. The detailed layer parameters are shown in Table~\ref{tab:fbs_para}. 
The voting module is the same as VoteNet. Note that, in training stage, we generate $M$ proposals from the votes by vote FPS (samples $M$ clusters based on votes’ XYZ), in test stage, we apply vote FPS on ScanNet V2 and seed FPS on SUN RGB-D (sample on seed XYZ and then find the votes corresponding to the sampled seeds).

The ray-based feature grouping module consists of two parts, \textit{Ray Point Generation} and \textit{Feature Enhancement by Determined Rays}. After generating a set of vote cluster centers $\{c_i\}_{i=1}^M$ based on the vote sampling and grouping, where $c_i = [v_i, f_i]$ (the vote center position $v_i \in \mathbb{R}^3$ and its corresponding features $f_i \in \mathbb{R}^{128}$), $M=256$ (the number of vote clusters). 
In our case, for each vote cluster, 66 rays are emitted uniformly from the cluster center with the determined angles and lengths generated in \S{\color{red}3.2.1}. We use a \textit{MLP [128, 128]} to regress the object scale of each cluster. 
As for the coarse-to-fine anchor point generation step, the number of coarse points ($K_c$) is 5 and fine points ($K_f$) is 3. To extract local feature of each anchor point, we apply two SA layers (coarse and fine), to aggregate the features of these seed points within a fixed radius (r = 0.2m) surrounding the query points. The two SA layers both have a receptive field specified by $r$=0.2m, a \textit{MLP[128, 64, 32]} for feature transform. But coarse layer samples 8 points by ball query operation and fine layer is 4 points. 
In term of the point mask prediction, we use a \textit{MLP[32+128, 32, 2]} to estimate the positive mask based on corresponding cluster features and local features. In \textit{Feature Enhancement module}, all the functions $\mathcal{F}^{**}$ are MLP networks. The detailed layer parameters are shown in Table \ref{tab:feh_para}.

The proposal module is a two-layer \textit{MLP[128, 128]}. We follow~\cite{qi2019deep} on how to estimate the 3D bounding boxes, except for size prediction that we adopts class-agnostic head to regress bounding box size directly. The layer’s output has \textit{5+2NH+3+NC} where the first five channels are for objectness classification and center regression (relative to the vote cluster center), \textit{2NH} channels are for heading bins classification and offsets regression, 3 is the scale regression for height, width and length, \textit{NC} is the number of semantic classes. In SUN RGB-D: \textit{NH} = 12, \textit{NC} = 10, and in ScanNet: \textit{NH} = 1, \textit{NC} = 18, due to the axis aligned bounding box. 

\subsection{RBGNet loss function details.}~As mentioned in the main paper, our model is trained end-to-end with a multi-task loss including foreground biased sampling $\mathcal{L}_\text{fbs}$, voting regression $\mathcal{L}_{\text{vote-reg}}$, ray-based feature grouping $\mathcal{L}_\text{rbfg}$, objectness $\mathcal{L}_\text{obj-cls}$, bounding box estimation $\mathcal{L}_\text{box}$, and semantic classification $\mathcal{L}_\text{sem-cls}$ losses.

\begin{equation}
    \begin{aligned}
    L = \lambda_\text{vote-reg}\mathcal{L}_{\text{vote-reg}} + \lambda_\text{fbs}\mathcal{L}_\text{fbs} + \lambda_\text{rbfg}\mathcal{L}_\text{rbfg} + \\\lambda_\text{obj-cls}\mathcal{L}_\text{obj-cls} + \lambda_\text{box}\mathcal{L}_\text{box} + \lambda_\text{sem-cls}\mathcal{L}_\text{sem-cls}.
    \end{aligned}
\end{equation}
Following the setting in VoteNet~\cite{qi2019deep}, we use the same loss terms $\mathcal{L}_{\text{vote-reg}}$, $\mathcal{L}_\text{obj-cls}$, $\mathcal{L}_\text{box}$ and $\mathcal{L}_\text{sem-cls}$, but $\mathcal{L}_\text{box}$ is class-agnostic and contains an additional corner loss defined in~\cite{qi2018frustum} for accurate bounding box estimation,

\begin{equation}
    \begin{aligned}
    \mathcal{L}_\text{box} = \lambda_\text{size-reg}\mathcal{L}_{\text{size-reg}} + \lambda_\text{corner}\mathcal{L}_\text{corner} + \\\lambda_\text{angle-cls}\mathcal{L}_\text{angle-cls} + \lambda_\text{angle-reg}\mathcal{L}_\text{angle-reg}.
    \end{aligned}
\end{equation}

As discussed in \S{\color{red}3.4}, $\mathcal{L}_\text{fbs}$ is a cross entropy loss used to supervise foreground sampling (see \S{\color{red}3.2}). 
$\mathcal{L}_\text{rbfg}$ is the sum loss of ray-based feature grouping module defined as follows:

\vspace{-10pt}
\begin{equation}
    \begin{aligned}
    \mathcal{L}_\text{rbfg} = \lambda_\text{scale-reg}\mathcal{L}_{\text{scale-reg}} +
    	 \lambda_\text{c-cls}\mathcal{L}_{\text{c-cls}} + \lambda_\text{f-cls}\mathcal{L}_{\text{f-cls}}.
    \end{aligned}
    \label{eq:loss}
\end{equation}
The balancing factors are set default as $\lambda_\text{vote-reg}\text{=}10.0$, $\lambda_\text{fbs}\text{=}3.0$, $\lambda_\text{rbfg}\text{=}10.0$, $\lambda_\text{obj-cls}\text{=}5.0$, $\lambda_\text{box}\text{=}10.0$, $\lambda_\text{sem-cls}\text{=}1.0$, $\lambda_\text{size-reg}\text{=}0.11$, $\lambda_\text{corner}\text{=}0.33$, $\lambda_\text{angle-cls}\text{=}0.1$, $\lambda_\text{angle-reg}\text{=}0.11$, $\lambda_\text{scale-reg}\text{=}0.11$, $\lambda_\text{c-cls}\text{=}0.2$ and $\lambda_\text{f-cls}\text{=}0.2$. 
$\mathcal{L}_{\text{scale-reg}}$, $\lambda_\text{size-reg}$ and $\lambda_\text{angle-reg}$ are all the smooth $\ell_1$ loss and their betas are 0.0625, 0.0625, 0.0400 separately.

\subsection{Other Grouping Mechanisms}
To further ablate the effectiveness of our ray-based feature grouping module, we refer several grouping strategies in 3D object detection as baselines and compare with them in \S{\color{red}4.4}. For a fair comparison, we only switch the feature aggregation mechanism while all other settings remain unchanged (\eg backbone with FBS, vote-FPS, proposal module). Here we give some detailed descriptions.

\noindent\textbf{Voting.}~The voting mechanism is first introduced by VoteNet~\cite{qi2019deep}. In our implementation, it is actually the VoteNet equipped with FBS, corner loss, vote-FPS in test stage and optimized hyperparameters.

\noindent\textbf{RoI-Pooling.}~For a given object proposal, the points within the predicted box are aggregated together. We adopt the similar implementation with \cite{cheng2021back}, predict the bounding boxes based on the voting cluster features and aggregate the points within the corresponding boxes by max-pooling. Finally, the aggregated features and cluster features are concatenated for 3D object detection. 

\noindent\textbf{Back-Tracing.}~It is first formulated in \cite{cheng2021back}. In our implementation, it is similar to RoI-Pooling described above, except that the prediction of bounding boxes is replaced by the maximum offsets of 6 directions. And then the points are aggregated by the balls uniformly sampled along the rays to enhance 3D bounding box estimation. 

\noindent\textbf{Group-free.}~We replace the ray-based feature grouping module with a transformer network adopted in Group-free~\cite{liu2021group}. For a fair comparison, we use vote-FPS for initial object candidate sampling instead of KPS. Then, the same as group-free~\cite{liu2021group}, we adopt the transformer as the decoder to leverage all the seed points to compute the object feature of each candidate. 

\section{Ray-based Representation Discussion} \label{sec:ray_discuss}
There are many choices for anchor point generation, such as classification~\cite{mildenhall2020nerf} and regression~\cite{xie2020polarmask, xu2019explicit}. 

\begin{table}
	\centering
	\resizebox{0.35\textwidth}{!}{
    	\setlength\tabcolsep{6pt}
		\renewcommand\arraystretch{1.0}
		\begin{tabular}{c|c|c}
			\hline\thickhline
			\rowcolor{mygray}
method         & mAP@0.25 & mAP@0.5 \\\hline
Reg-Ray(R66)            & 68.0  & 50.2 \\
Our(R66)            & \textbf{69.6}  & \textbf{53.6}\\

\hline
\hline

	\end{tabular}
	}
	\caption{The average performance of different ray representations on ScanNet V2.
	}
	\label{tab:regray}
	
\end{table}

\begin{figure*}
  \centering
  \vspace{-10pt}
   \includegraphics[width=1.0\linewidth]{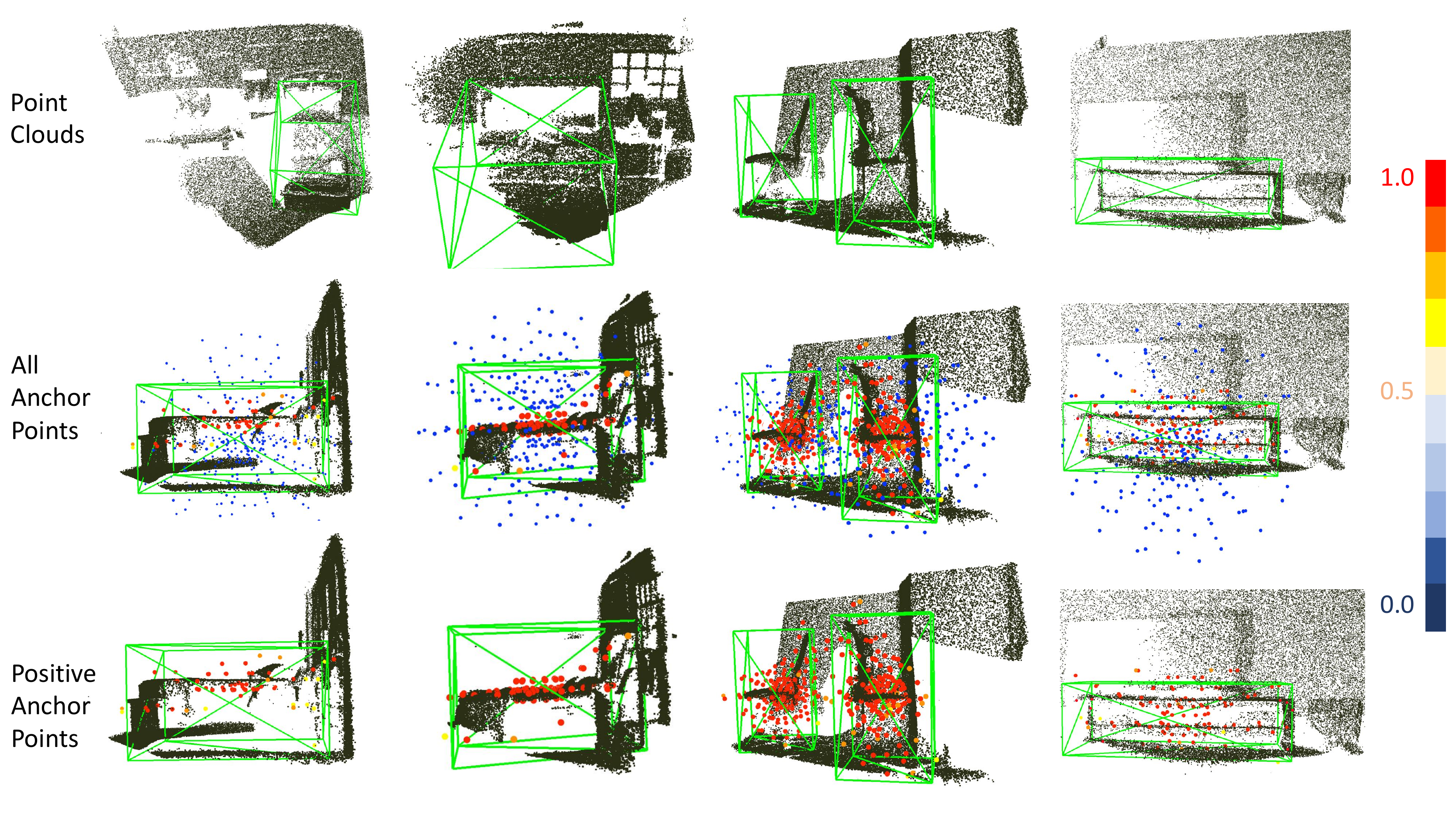}
   \caption{Qualitative results of shape distribution our model learned.}
   \label{fig:shapescore}
\end{figure*}

\begin{itemize}[leftmargin=*]
	\setlength{\itemsep}{0pt}
	\setlength{\parsep}{-2pt}
	\setlength{\parskip}{-0pt}
    \setlength{\leftmargin}{-10pt}
	
	\item \textbf{Regression:} Given the center and surface points of an object, they want to represent the shape by polar coordinates. The length of $n$ rays can be computed easily. Then, the model regresses the length of each ray and captures the points when the ray terminates somewhere. 
	\item \textbf{Classification:} Given an instance, model predicts far bounds of all rays and samples a fixed number of potential query points on each ray, and then extracts local features and classifies those points whether belong to corresponding object to generate reasonable anchor points. Our model adopts this way and we will discuss why do we choose it.
\end{itemize}
In 2D perception community, some methods also represent object shape by rays~\cite{xie2020polarmask, xu2019explicit} in regression way, which applies the angle and distance as the coordinate to locate points. 
However, due to the particular property of point clouds, this regression pipeline has many problems in 3D scenario, \ie, i) center is outside of the object, no intersection with the object surface at some angles, ii) limited expressive ability on concave shape, one ray may has multiple intersections. Compare to regression, classification pipeline is more reasonable to represent point clouds and doesn't have the above problems, so we choose it to generate anchor points. In Table~\ref{tab:regray}, we show the results of the two representations, classification approach performs better than regression. 

\section{More Results}\label{sec:more}
\subsection{Per-class Evaluation} \label{sec:perclass}
We evaluate per-category on ScanNet V2 and SUN RGB-D under different IoU thresholds. Table~\ref{tab:scan0.25} and Table~\ref{tab:scan0.5} report the results on 18 classes of ScanNetV2 with 0.25 and 0.5 box IoU thresholds respectively. Table~\ref{tab:sun0.25} and Table~\ref{tab:sun0.5} show the results on 10 classes of SUN RGB-D with 0.25 and 0.5 box IoU thresholds. Our approach outperforms the baseline VoteNet~\cite{qi2019deep} and prior state-of-the-art methods Group-free~\cite{liu2021group} significantly in almost every category. These improvements are achieved by using ray-based feature grouping and foreground biased sampling to better encode object surface geometry.

\subsection{Visualization of Positive Anchor Points} \label{sec:shapescore}
Fig.\ref{fig:shapescore} shows the scores of coarse anchor points predicted from our RBGNet in a typical SUN RGB-D scene. We clearly see that the high responses are almost on the object surface (bed, chair \etc) while low responses are on the empty space or background surface. This verifies that our method can really learn the shape distribution and boost point-based 3D detectors.

\subsection{Quantitative Results} \label{sec:qresults}
\begin{figure*}[t]
  \centering
  \vspace{-10pt}
   \includegraphics[width=0.88\linewidth]{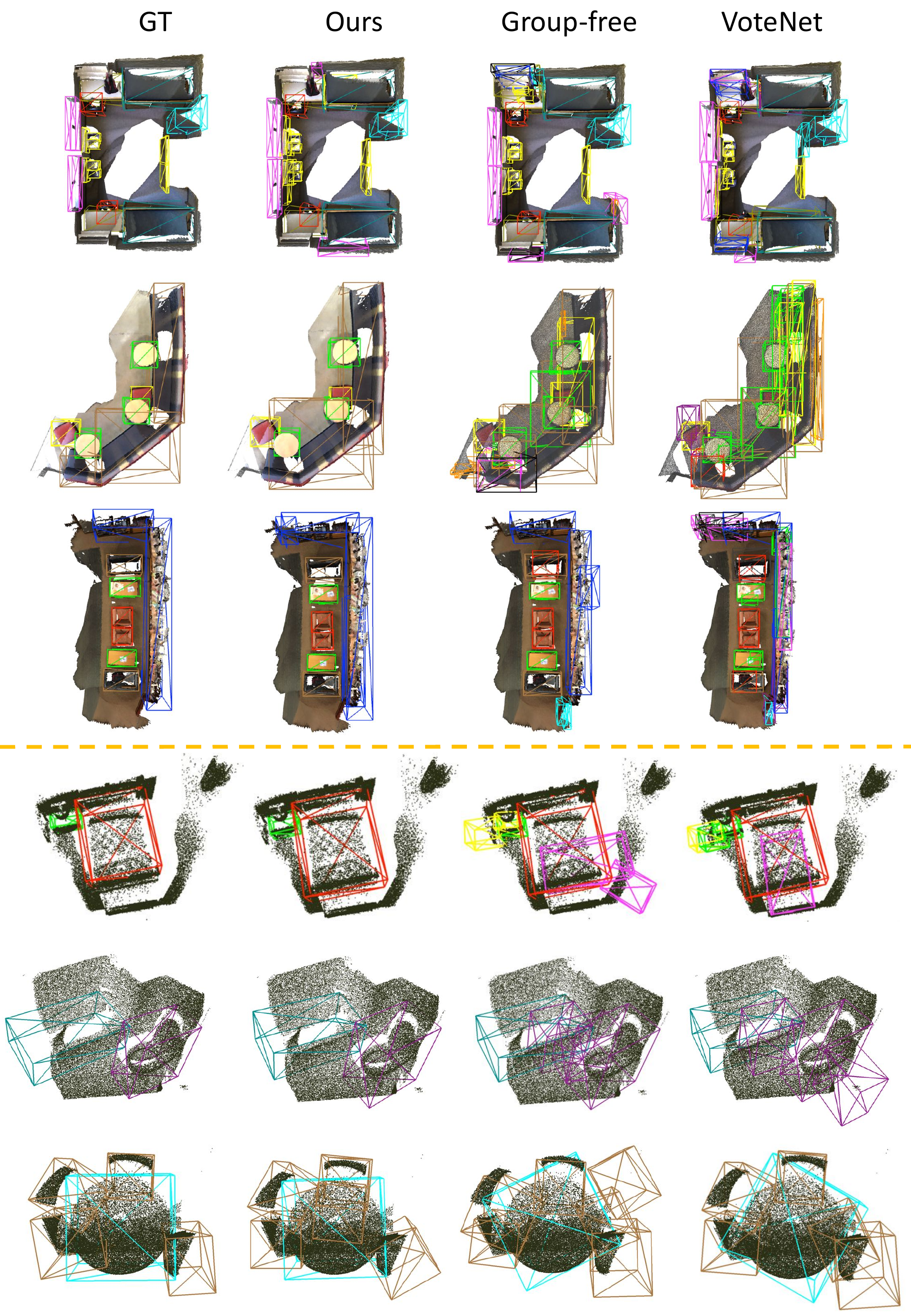}
   \caption{Qualitative results on ScanNet V2(top) and SUN RGB-D(down). The baseline methods are Group-free~\cite{liu2021group} and VoteNet~\cite{qi2019deep}.}
   \label{fig:qualitative_results}
\end{figure*}

We provide more qualitative comparisons between our method and the top-performing reference methods, such as Group-free~\cite{liu2021group} and VoteNet~\cite{qi2019deep}, on the ScanNet V2 and SUN RGB-D datasets. Please see Fig.~\ref{fig:qualitative_results} for more qualitative results.

\section{Limitations}\label{sec:discuss}
Although our method achieves promising performance on multiple datasets, there are still some limitations. Compared with the previous approaches, the performance of RBGNet is significantly better in the case of a large number of rays. However, there is a trade-off between computational cost and performance improvement, as shown in main paper. In the future, we hope to discover approaches that can encode surface geometry more efficiently.  

\end{document}